\PassOptionsToPackage{table}{xcolor}
\documentclass{article}

\PassOptionsToPackage{numbers,compress}{natbib}
\usepackage{neurips_2025}

\usepackage{caption}   
\usepackage{setspace}  

\usepackage{wrapfig}
\usepackage[utf8]{inputenc} 
\usepackage[T1]{fontenc}    
\usepackage{hyperref}       
\usepackage{url}            
\usepackage{booktabs}       
\usepackage{amsfonts}       
\usepackage{nicefrac}       
\usepackage{microtype}      
\usepackage{xcolor}         
\usepackage{graphicx} 
\usepackage{amsmath} 

\usepackage[ruled,vlined]{algorithm2e}
\usepackage[numbers]{natbib}
\usepackage{array}

\usepackage{amssymb}  
\usepackage{booktabs}
\usepackage{pifont}  

\usepackage[table]{xcolor}
\usepackage{colortbl}
\usepackage{subcaption}                
\usepackage{multirow}

\newcolumntype{C}{>{\centering\arraybackslash}S}
\newcommand{\xmark}{\ding{55}} 

\usepackage{siunitx}
\usepackage{enumitem}   
\usepackage[utf8]{inputenc}  
\usepackage[T1]{fontenc}     

\title{NSW-EPNews: A News-Augmented Benchmark for Electricity Price Forecasting with LLMs}

\author{
  Zhaoge~Bi \\ The University of Sydney \\ \texttt{zhbi4108@uni.sydney.edu.au}
  \And
  Linghan~Huang \\ The University of Sydney \\ \texttt{lhua5130@uni.sydney.edu.au}
  \And
  Haolin~Jin \\ The University of Sydney \\ \texttt{hjin3177@uni.sydney.edu.au}
  \And
  Qingwen~Zeng \\ The University of Sydney \\ \texttt{qzen5227@uni.sydney.edu.au}
  \And
  Huaming~Chen \\ The University of Sydney \\ \texttt{huaming.chen@sydney.edu.au}
}

\begin{document}
\maketitle

\begin{abstract}

    Electricity price forecasting is a critical component of modern energy‐management systems, yet existing approaches heavily rely on numerical histories and ignore contemporaneous textual signals. We introduce NSW-EPNews, the first benchmark that jointly evaluates time-series models and large language models (LLMs) on real-world electricity-price prediction. The dataset includes over 175,000 half-hourly spot prices from New South Wales, Australia (2015–2024), daily temperature readings, and curated market‐news summaries from WattClarity. We frame the task as 48-step-ahead forecasting, using multimodal input, including lagged prices, vectorized news and weather features for classical models, and prompt-engineered structured contexts for LLMs. Our datasets yields 3.6k multimodal prompt-output pairs for LLM evaluation using specific templates. Through compresive benchmark design, we identify that for traditional statistical and machine learning models, the benefits gain is marginal from news feature. For state-of-the-art LLMs, such as GPT-4o and Gemini 1.5 Pro, we observe modest performance increase while it also produce frequent hallucinations such as fabricated and malformed price sequences. NSW-EPNews provides a rigorous testbed for evaluating grounded numerical reasoning in multimodal settings, and highlights a critical gap between current LLM capabilities and the demands of high-stakes energy forecasting.  

\end{abstract}
\section{Introduction}
\label{sec:intro}
Time-series forecasting is central to modern infrastructure, and electricity-price prediction is its canonical, high-stakes application. Accurate and transparent price outlooks steer load balancing, dispatch planning, and wholesale trading across deregulated markets \cite{cornell2023_intro_probabilistic_NEM}. Classical econometric tools—ARIMA and its seasonal extension SARIMA—remain popular thanks to their interpretability, yet their linear assumptions limit fidelity to long-range, nonlinear structure in prices \cite{box1994ARIMA,mancha2024_intro_ARIMA_limits}. Conventional machine-learning regressors (LR, RF, SVR) capture richer covariate effects, but still falter when the signal is buried in complex temporal dependencies or interleaved with unstructured context such as news headlines \cite{balli2021data_intro_ML_forecast,ahmed2021_intro_ml_fNews_detect,junque2014_intro_SVR_textunderstand}. In contrast, transformer-based large language models (LLMs) have recently demonstrated strong zero-/few-shot competence on time-series tasks by treating numeric sequences as another modality of “language” and conditioning on heterogeneous side information \cite{liu2025_TimeCMA}. Their capacity to ingest both structured features and free-form text—e.g., market commentary and temperature forecasts—raises the prospect of materially improved electricity-price prediction.

\paragraph{Motivation.}
Electricity prices are driven by a confluence of factors that extend far beyond historical values. Market commentary and policy news shape expectations and trading behaviour \cite{lazarczyk2016MarketNews,rogmann2024MediaSentiment,wei2024NewsPropaganda}. Weather also exerts a first-order effect: extreme temperatures trigger spikes in cooling or heating demand, propagating into price volatility \cite{dubey2021_liter_SARI_LSTM_study,mosquera2024weather}. Harnessing such exogenous signals therefore promises improved forecast accuracy; When information from multiple sources—such as weather data and news content—is integrated, the natural language processing capabilities of LLMs become particularly advantageous.


\paragraph{Gap.}
In recent years, time-series forecasting evaluation has coalesced around public benchmarks—spanning traffic, energy load, macro-economic, and weather datasets—with error metrics like MSE and MAE (and variants) plus significance testing to compare models \cite{zeng2022transformerseffectivetimeseries,Lago_2021}. In electricity-price forecasting, open datasets covering multiple regional markets and year-long test sets against state-of-the-art baselines have become standard \cite{Lago_2021}. Yet these benchmarks overlook unstructured news data, despite evidence that external text can enhance numerical prediction \cite{andrei2024energypricemodellingcomparative}. With large language models (LLMs), zero-shot forecasting and multimodal fusion show great promise—for example, GPT4MTS’s integration of news events and numerical series yields significant accuracy gains \cite{10.1609/aaai.v38i21.30383}. To advance the field, a benchmark that fuses electricity-price time series with news text is both novel and necessary—but must be paired with rigorous prompt engineering and fine-tuning to mitigate LLM hallucination risks \cite{jin2024positionlargelanguagemodels}. Despite the intuitive appeal of text-enriched forecasting, the community lacks a systematic benchmark for assessing how well LLMs integrate unstructured information into price predictions. Whether LLMs can reliably translate news sentiment and weather cues into numerical forecasts—\emph{and whether their outputs can be trusted in high-stakes markets}—remains an open question \cite{liu2025_TimeCMA,chen2024_intro_chatgptStock}.

Our contribution can be summarized as follows:
\begin{itemize}
    \item \textbf{Dataset Construction:} We introduce NSW-EPNews, a large-scale multimodal dataset and benchmark for electricity price forecasting that integrates historical electricity price time series, meteorological data (temperature records), and relevant news articles from New South Wales (NSW, Australia) covering 2015–2024. This dataset provides a comprehensive testbed for analyzing how rich multimodal inputs (weather and news) influence electricity price dynamics over nearly a decade of real-world data. \textbf{Data Availability:} We have fully released our benchmark dataset, experimental process, experimental results, which is now available at the following link: \url{https://figshare.com/s/e25f3a98679d347f2a2e}. 

    \item \textbf{Prompt Template Design:} We design a prompting methodology that enables both classical time-series forecasting models and large language models (LLMs) to ingest and make predictions from the combined multimodal data. In particular, we craft input templates that present the numerical (price and weather) data and textual news information in a unified format, allowing models of different types to effectively utilize the heterogeneous inputs and enabling direct comparison of traditional forecasting approaches against LLM-based methods on the same tasks.

    \item \textbf{Evaluation Framework:} We develop an evaluation framework to rigorously assess model performance, quantifying both standard forecasting accuracy and hallucination-related failure modes. This framework computes conventional prediction error metrics for price forecasts while also detecting and measuring any hallucinated content in model outputs, thereby evaluating each model’s predictive accuracy alongside the factual consistency and reliability of its generated explanations or reports.

    \item \textbf{Key Findings:} Extensive experiments show that (i) traditional statistical and machine-learning baselines gain only marginal benefit from naïvely vectorised news, and (ii) state-of-the-art LLMs incur large errors (MAE~$>$ 100 AUD/MWh on some splits) and frequent hallucinations.  Current prompting and model architectures are therefore insufficient for robust, news-aware electricity-price forecasting, underscoring the need for improved prompt engineering, retrieval grounding and fine-tuning strategies.
\end{itemize}

\section{Background}
\paragraph{Traditional Time Series Forecasting vs LLMs.}
Traditional time-series forecasting has long been dominated by statistical models and machine learning methods. Classical models like ARIMA rely on assumptions of linearity, stationarity, and carefully tuned seasonal heuristics\cite{jin2024positionlargelanguagemodels}. These models perform well for small-scale data with stable patterns but struggle as data complexity grows. In recent years, deep learning approaches (e.g. LSTMs and Transformers) have become state-of-the-art on many forecasting benchmarks, demonstrating the ability to learn non-linear and long-term dependencies directly from data
\cite{yang2024timeragboostingllmtime}. Yet even advanced deep models often require structured inputs and feature engineering, making it difficult to incorporate unstructured information (such as free-form text or images). They can still miss hidden complex patterns in large-scale, diverse sequences
\cite{yang2024timeragboostingllmtime}, highlighting an opportunity for more flexible frameworks. Large Language Models (LLMs) have emerged as a promising new avenue for forecasting, inspired by their success in capturing complex patterns in natural language. LLM-based models leverage extensive pre-training on vast corpora to integrate broader contextual knowledge and act as general pattern learners beyond just language\cite{andrei2024energypricemodellingcomparative}. Initial studies have only begun exploring LLMs for time-series tasks; for example, TimeGPT and other LLM forecasters have demonstrated competitive accuracy in certain conditions, though their advantages vary with market volatility. Notably, the energy domain presents an open opportunity – until recently no LLM had been applied to electricity price forecasting, motivating investigation into whether such models can be adapted to improve predictive performance\cite{andrei2024energypricemodellingcomparative}. By harnessing pre-trained knowledge and zero-shot capabilities, LLM-based forecasting methods aim to overcome the rigidity of traditional models, especially in handling heterogeneous and unstructured data sources.

\paragraph{External Influencing Factors in Electricity Price Forecasting.}
Electricity prices are driven by a multitude of external factors beyond the historical price series itself. Weather conditions, especially temperature, have a well-documented impact on electricity demand and thus prices. Prior studies show that temperature alone can explain a large portion of demand variability – for instance, temperature changes accounted for approximately 75\% of short-term load variation in one analysis\cite{4414354}. This strong relationship means that accurate temperature forecasts can significantly enhance price prediction accuracy, as demand-driven price fluctuations are better anticipated when weather dynamics are taken into account. In addition to meteorological factors, informational and socio-economic factors play a critical role. News events, policy announcements, and market sentiment can all sway trader expectations and induce price movements. Recent research has demonstrated the value of incorporating textual data: for example, leveraging GPT-based analysis of energy news can produce features that align closely with subsequent price fluctuations\cite{en17102338}. Such findings indicate that news articles and expert reports often contain leading indicators of market dynamics, from geopolitical developments to technical disruptions. By converting unstructured inputs (like news or social media sentiment) into quantitative signals\cite{en17102338}, forecasters can capture effects that traditional time-series models would otherwise miss. Overall, the evidence is clear that external influencing factors – whether exogenous physical drivers like weather or informational drivers like news – materially affect electricity prices, underscoring the importance of models capable of ingesting these heterogeneous data sources.

\paragraph{Hallucination and Prompt Engineering.}
LLM-based forecasters frequently generate implausible trajectories—spikes, offsets, or flat lines—that are unsupported by the input data, compromising trust in high-stakes settings \citep{bang2023multitaskmultilingualmultimodalevaluation}. Recent work mitigates this by grounding the prompt with retrieved, similar subsequences: retrieval-augmented generation (RAG) constrains outputs to historical patterns and demonstrably curbs hallucinations \citep{lewis2021retrievalaugmentedgenerationknowledgeintensivenlp}. Complementary progress comes from prompt engineering. Zero-shot instructions test a model’s native generalisation, whereas few-shot templates supply a handful of input–output exemplars, yielding sizeable accuracy gains via in-context learning \citep{brown2020languagemodelsfewshotlearners}. Further, chain-of-thought prompting elicits intermediate reasoning steps that boost multi-step prediction quality \citep{wei2023chainofthoughtpromptingelicitsreasoning}. Together, retrieval grounding and carefully designed prompts represent the current best practice for deploying LLMs in time-series forecasting while keeping hallucinations in check.

\section{NSW-EPNews Benchmark Design}

\subsection{Data Preprocess}
\paragraph{Scrape news.}
We scraped ten years of news articles gathered from WattClarity\cite{wattclarity2025wattclarity}. WattClarity is established in 2007, it provides detailed analysis and regularly publishes news related to the operation of the Australian NEM, making it a trusted and authoritative source of electricity market information. To illustrate the practical relevance of integrating news into forecasting, we highlight several real scenarios where electricity market news significantly influenced price dynamics (See appendix \ref{app:News article}). Below is an algorithm that demonstrates how our scrape script works.

\begin{algorithm}[H]
  
\caption{News scraping process}

\KwIn{\texttt{start\_url}}
current$\!\leftarrow\!$\texttt{start\_url};\;
\While(\tcp*[f]{crawl archive}){(soup$\!\leftarrow\!$\texttt{get\_soup}(current))$\neq$\texttt{null}}{
  \lForEach{link $\in$ \texttt{fetch\_news\_links\_from\_page}(soup)}%
    {\If{(art$\leftarrow$\texttt{get\_soup}(link))$\neq$\texttt{null}}%
      {(title,author,raw\_date,date,topic)$\leftarrow$\texttt{extract\_information}(news);\,
       content$\leftarrow$\texttt{classify\_content}(\texttt{extract\_content}(news));\,
       \texttt{append\_row\_to\_csv}(title,author,date,topic,content);}}%
  current$\!\leftarrow\!$\texttt{get\_next\_page\_url}(soup);\,
  \texttt{sleep}(shortInterval);}
\end{algorithm}


\paragraph{Classify news.}
Electricity-market reports are often long and heterogeneous; directly feeding the raw text into a forecasting model obscures which portions of the narrative actually matter for prices. To convert these articles into machine-parsable signals we decompose the instruction to GPT-4o into four modular blocks, each targeting a specific weakness observed in baseline LLM processing. Applying the four-block prompt,including Role assignment, Classification criteria, Key attributes, Summary rules to the entire corpus yields a concise paragraph plus structured metadata for each of the $\sim$3.6 k articles. This dual summarisation–annotation step serves two purposes: (i) it \emph{compresses} lengthy documents so that LLM forecasters can attend to salient facts without exceeding context limits, and (ii) it \emph{labels} every sample with an impact level, allowing us to probe whether models differentially exploit high-impact versus low-impact news.(More details in appendix F)


\begin{figure}[htbp]
  \centering
  \includegraphics[width=1\linewidth,page=1]{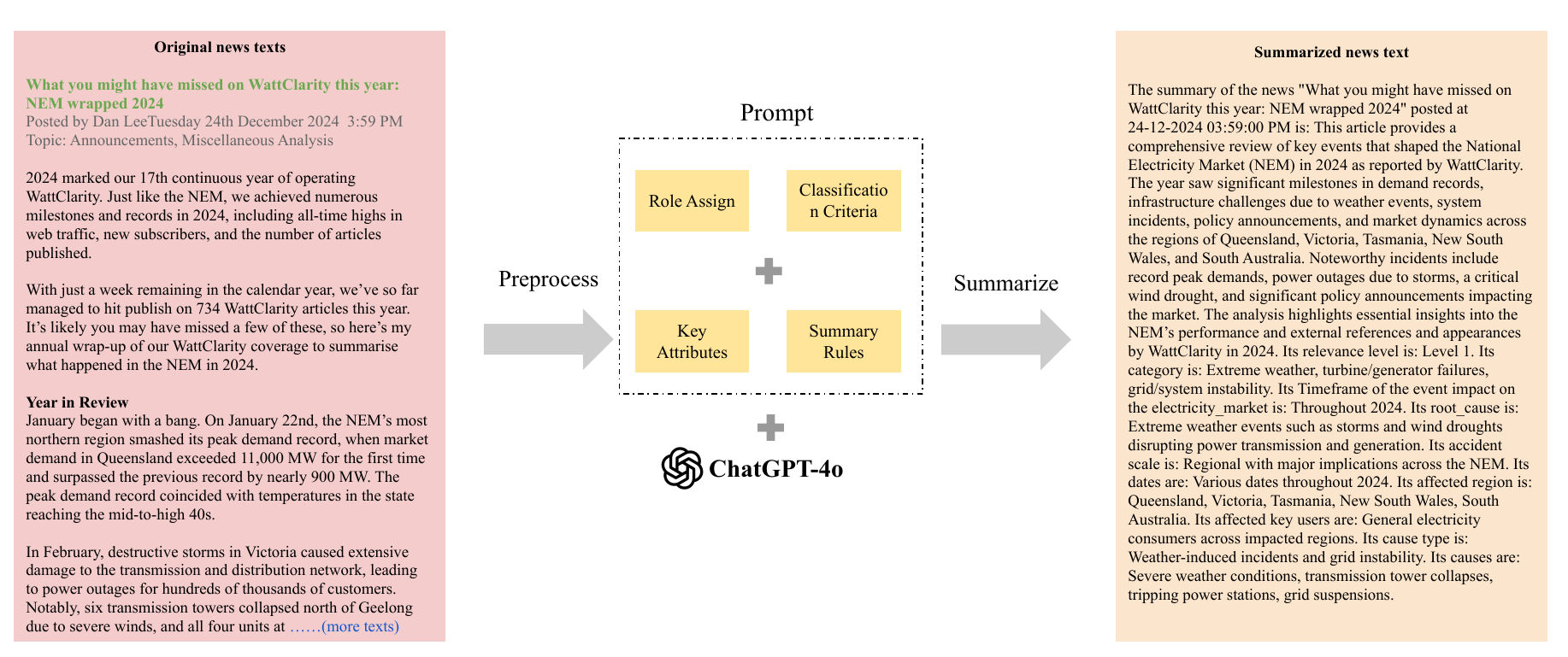}
  \caption{Prompt used for news classification}
  \label{fig:example}
\end{figure}

\paragraph{Price data preprocess.}
Our benchmark includes ten years of Australia New South Whales' electricity price data collected from the Australian National Electricity Market (NEM)\cite{aemo2025nem}. The data frequency of electricity price data recording were thirty-minutes until 1st October 2021, NEM changed the frequency to five-minutes. To ensure consistency across the entire dataset, we applied a down-sampling method to the post-2021 data. This method aggregats every six 5-minute records into a single thirty-minute interval using the median value to preserve the central trend while reducing the influence of short-term fluctuations.

\begin{figure}[t]                
  \centering
  \begin{minipage}[t]{0.48\linewidth}
    \scriptsize                                   
    \setstretch{0.92}   
    \captionsetup{type=algocf}
    \begin{algorithm*}[H]                         
      \DontPrintSemicolon
      \captionof{algocf}{Downsample 5$\!\to\!$30-min electricity prices}
      \KwIn{records sorted by timestamp}
      \KwOut{30-minute data}
      cutoff $\leftarrow$ \texttt{2021-10-01 00:00}\;
      output $\leftarrow$ {[}r for r in records if r.time $<$ cutoff{]}\;
      post   $\leftarrow$ {[}r for r in records if r.time $\ge$ cutoff{]}\;
      \For{window of each 6 consecutive r in post}{
        $t_0\!\leftarrow\!$ window[0].time\;
        $v\!\leftarrow\!\texttt{median}(\{r.value\})$\;
        output.append(($t_0$, $v$))\;
      }
      \Return output\;
    \end{algorithm*}
  \end{minipage}
  \hfill                                 
  \begin{minipage}[t]{0.48\linewidth}
    \scriptsize
    \setstretch{0.92}
    \captionsetup{type=algocf}
    \begin{algorithm*}[H]
      \DontPrintSemicolon
      \captionof{algocf}{Simplified forecasting pipeline}
      \KwIn{Prices, classified news, temperature}
      \KwOut{Baseline metrics, LLM metrics}
      prices $\leftarrow$ \texttt{load("prices")}\;
      news   $\leftarrow$ \texttt{load("news")}\;
      temps  $\leftarrow$ \texttt{load("temps")}\;
      raw $\leftarrow$ [ ]\;
      \ForEach{day}{
        raw.append(\{prices: prices[day],
        news: \texttt{summarize}(news,day),
        temp: \texttt{stats}(temps,day)\})\;
      }
      \ForEach{model $\in$ \{ARIMA, XGB, LR\}}{
        baseline[model] $\leftarrow$ \texttt{train\_eval}(model, raw)\;
      }
      prompts $\leftarrow$ \texttt{format\_prompts}(raw)\;
      \ForEach{p $\in$ prompts}{
        llm\_results.append(\texttt{call\_llm}(p))\;
      }
      \Return \{baseline, llm\_results\}\;
    \end{algorithm*}
  \end{minipage}
\end{figure}

\subsection{Data Processing}

\paragraph{Raw Dataset construction.}
Records in raw dataset are stored in JSON format.  Each record contains four key cmoponents: (1) the electricity price data for a single day (48 half-hourly data points), (2) the summarized news articles posted on that same day, (3) The maximum and minimum temperature, and (4) the electricity price data for the following day (also 48 data points), which serves as the ground truth for forecasting. By doing this, each record contains one day of historical electricity price data and the true prices for the following day as the prediction target. The next record uses the previous day's true prices as its historical input. This creates a sliding window structure where one day of data is used to forecast the next day, allowing continuous day-by-day evaluation. 

\begin{wrapfigure}{r}{0.46\linewidth}  
  \vspace{-\intextsep}                 
  \centering
  \includegraphics[width=\linewidth,page=1]{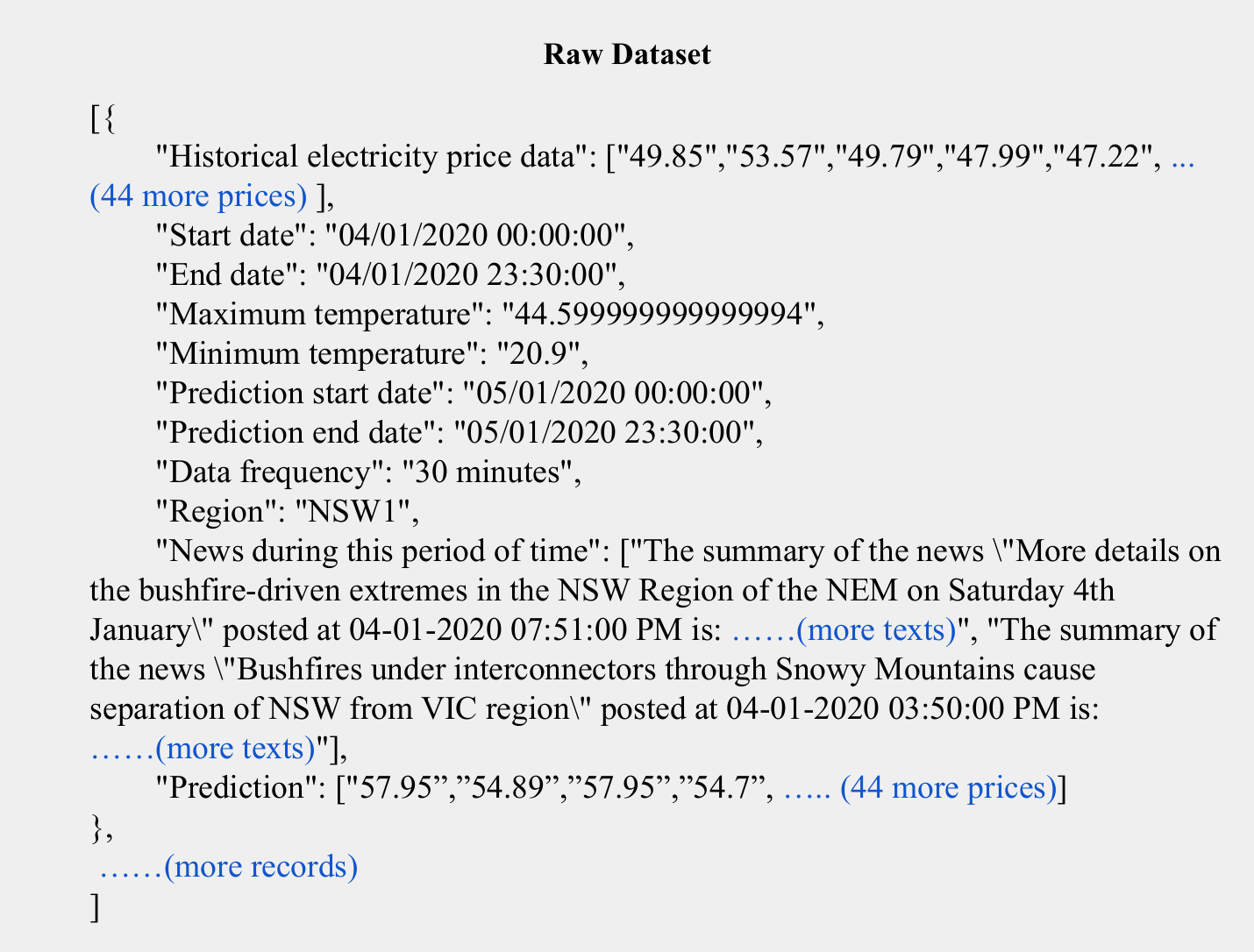}
  \caption{Example record in the raw dataset}
  \label{fig:raw-example}
\end{wrapfigure}

The raw dataset will eventually have more than 3600 pairs of records. It primarily serves two purposes: testing the performance of baseline forecasting models and constructing prompt–target pairs for LLM-based evaluation. There is an example record shown in Figure.~\ref{fig:raw-example}. This raw dataset can be used as input for statistical and traditional machine learning models. Statiscial models use the numerical price data only. Vectorized news content and temperature data can be used as features for machine learning methods. These baseline models generate predicted electricity prices, which are then directly compared against the ground truth values to evaluate forecasting accuracy.


\paragraph{LLM Prompt Construction.}

To test the performance of LLMs, we further preprocess the raw dataset into a new format. In this dataset, each record includes a constructed prompt and a corresponding sequence of true prices as the prediction targets, forming a complete input–output pair for evaluating the LLM’s forecasting performance. Each pair is still saved in JSON format, same as raw dataset. This setup allows for a direct comparison between the model’s predicted results and the ground truth values using standard evaluation metrics. It follows the same sliding window setting as the raw dataset, using one day of historical price data and background information to predict the next day’s prices. Although LLMs are highly capable of processing natural human language, they can also suffer from issues such as hallucination\cite{ji2023HallucinationSurvey}, generating outputs that are fluent but factually incorrect. A single prompt format is insufficient to reliably guide LLMs toward producing accurate and reasonable forecasts\cite{mesko2023promptEngMed}. Therefore, in our benchmark, we design and evaluate four different prompt formats. These formats vary in two key dimensions: whether they are zero-shot or few-shot, and whether they include a chain-of-thought\cite{wei2023COT} reasoning process or not. The information in the raw dataset was extracted and reorganized to create four groups of distinct datasets, each corresponding to one of the four prompt styles (See appendix \ref{app:promtsettings}): zero-shot, zero-shot with chain-of-thought, few-shot, and few-shot with chain-of-thought. The aim is to apply these prompting techniques to enhance the prediction performance of LLMs and to explore how their behavior varies across different prompt settings. This allows us to analyze the models’ ability to incorporate news information into forecasting from multiple perspectives. LLMs are expected to receive prompts that integrate 48 historical price points from a single day, corresponding news summaries, temperature readings, and date information, and to output a sequence of 48 forecasted price values for the following day in the format we have defined. To evaluate the performance of LLMs under various conditions by using both standard evaluation metrics and hallucination detection methods defined by ourselves. To better illustrate the effects of prompt engineering and background information, our benchmark also includes a simplified prompt-target dataset for ablation study. This version's prompt follows the zero-shot style and contains only historical electricity prices and datetime information in the prompt, allowing the LLMs to make predictions without relying on news or temperature data. For more details about the different prompt settings, please see the appendix A.

\begin{table}[h]

\centering
\caption{Prompt Format Comparison Across Input Features and Design Strategies}
\begin{tabular}{lcccccc}
\toprule
\textbf{Prompt Type} & \textbf{Price History} & \textbf{News} & \textbf{Temperature Data} & \textbf{Q\&A} & \textbf{CoT Reasoning} \\
\midrule
Zero-shot          & \checkmark & \checkmark & \checkmark & \xmark & \xmark \\
Few-shot           & \checkmark & \checkmark & \checkmark & \checkmark & \xmark \\
Zero-shot + CoT    & \checkmark & \checkmark & \checkmark & \xmark & \checkmark \\
Few-shot + CoT     & \checkmark & \checkmark & \checkmark & \checkmark & \checkmark \\
Ablation Study     & \checkmark & \xmark     & \xmark     & \xmark     & \xmark \\
\bottomrule
\end{tabular}
\end{table}

\subsection{LLMs Hallucinations and Error Detection}
We introduced supplementary measures to assess the reasoning quality of LLM-generated forecasts. Since LLMs are known to suffer from hallucinations and false responses, we observed several instances where their outputs were illogical or inconsistent with the given inputs. These behaviors indicate that the model may not be reasoning about future prices based on a true understanding of the provided structured and unstructured information. To detect such cases, we implemented additional checks during evaluation (See Algorithm 3 below). Specifically, we compare the LLM's predictions with the historical input data and record the number of occurrences where direct copying, offset-based generation, or repetition patterns are observed. The rate of these hallucinations and errors is calculated by dividing the number of occurrences by the total number of prompts processed by the LLMs. In our benchmark, we are trying to detect four different types of hallucination and error outputs. They are \textbf{echoing failure}, \textbf{trival transformation}, \textbf{degenerate copy} and \textbf{format violation} (please see appendix \ref{app:hallucination}).

\begin{algorithm}[ht]
\small  
\DontPrintSemicolon                
\KwIn{history, K}
\KwOut{format\_violation, echoing\_failure, trivial\_transformation, degenerate\_copying}

\ForEach{each LLM API call}{
  Try to parse the model’s reply into a list {\ttfamily pred}\;
  \If{parsing fails}{
    {\ttfamily format\_violation} $\leftarrow$ true\;
    \textbf{continue}\;
  }
  $\textit{echo\_count}\leftarrow$ \#values in {\ttfamily pred} matching any in {\ttfamily history}\;
  \If{$\textit{echo\_count}\ge10$}{
    {\ttfamily echoing\_failure}$\leftarrow$ true\;
  }
  $\textit{offsets}\leftarrow\{\;\text{pred}[i]-\text{history}[i]\mid i=1..K\}\;$\;
  \ForEach{offset in offsets}{
    $\textit{match\_count}\leftarrow\#\{j\mid\text{pred}[j]=\text{history}[j]+offset\}$\;
    \If{$\textit{match\_count}\ge20$}{
      {\ttfamily trivial\_transformation}$\leftarrow$ true\;
      \textbf{break}\;
    }
  }
  $\textit{freq\_map}\leftarrow$ frequency map of values in {\ttfamily pred}\;
  \If{any value’s freq $>5$}{
    {\ttfamily degenerate\_copying}$\leftarrow$ true\;
  }
  Record the four boolean flags for this API call\;
}
\caption{Hallucination and error output detection pseudocode}
\end{algorithm}

\section{Results}


\paragraph{Standard Evaluation Metrics.}
We evaluate forecasting performance using four standard metrics: MSE, RMSE, MAE and MAPE. To account for irregular output lengths from LLMs, any extra predicted values are discarded, and missing values are temporarily filled with nulls and excluded from metric calculations . MSE highlights large errors by squaring deviations, making it sensitive to outliers. RMSE is the square root of MSE, expressed in the same unit as the original data, enhancing interpretability. MAE captures the average absolute error in dollars and is less affected by extreme values. MAPE measures relative error as a percentage but is unstable when true prices are near zero, which is common in the Australian market; thus, zero-value entries are excluded when computing MAPE (See appendix \ref{app:Evaluationmetrics} for more details).

\paragraph{Model Used.} 
The baseline models include the classical statistical model ARIMA and two machine learning approaches: Linear Regression (LR) and XGBoost. ARIMA serves as a fully numerical baseline that models only the autoregressive structure of the price series, while LR and XGBoost incorporate multimodal features, including recent prices, TF-IDF-encoded news, and temperature data. These models are trained and evaluated on multiple temporal splits of the dataset (full, 50\%, 30\%, and 10\%) and provide reference points for comparison with large language models (LLMs). The selected LLMs, ChatGPT-4o and Gemini 1.5 Pro, are tested using four prompt engineering strategies: zero shot, few shot, zero shot with chain-of-thought (CoT), and few shot with CoT. Each model is evaluated on the same dataset splits. Prompts are submitted through iterative API calls, and generated outputs are evaluated for forecasting accuracy using MAE, MSE, RMSE, and MAPE. Additionally, outputs are examined for hallucinations, including format violations, direct copying of historical prices, trivial transformations such as constant offsets, and degenerate copying where a single value is repeated throughout. This comprehensive evaluation provides a clear comparison between traditional forecasting models and modern LLM-based approaches. (Experiment settings: appendix \ref{app:experimentsettings})


\subsection{Baslines and LLMs results}

\begin{table}[ht]
\centering
\resizebox{1\textwidth}{!}{%
\begin{tabular}{
  ll
  l       
  S[table-format=7.2]       
  S[table-format=4.2]       
  S[table-format=4.2]       
  S[table-format=5.2]       
  l     
  S[table-format=1.2]       
  C[table-format=1.4]       
  l      
  S[table-format=1.2]       
  C[table-format=1.4]       
  l       
  S[table-format=1.2]       
  C[table-format=1.4]       
  l       
  S[table-format=1.2]       
  C[table-format=1.4]       
}
\toprule
Model & Prompt 
  & \multicolumn{4}{c}{Recent 50\% of the dataset} 
  & \multicolumn{4}{c}{Recent 30\% of the dataset} 
  & \multicolumn{4}{c}{Recent 10\% of the dataset} \\
\cmidrule(lr){3-6}\cmidrule(lr){7-10}\cmidrule(lr){11-14}
 & 
  & {MSE} & {RMSE} & {MAE} & {MAPE/\%} 
  & {MSE} & {RMSE} & {MAE} & {MAPE/\%} 
  & {MSE} & {RMSE} & {MAE} & {MAPE/\%} \\
\midrule
ARIMA              & None         & 
  124926.234 & 353.449 &  54.579 & 1041.600 
& 158787.618 & 398.482 &  66.755 & 1612.790 
& 393097.454 & 626.975 &  89.092 & 1714.130 \\
Linear Regression  & None         & 
  129536.826 & 359.912 &  69.537 &  812.475 
& 165339.979 & 406.620 &  85.795 & 1328.577 
& 393662.787 & 627.426 &  90.195 & 1599.455 \\
XGBoost            & None         & 
  263069.003 & 512.903 & \textcolor{red}{113.330} & 1149.020 
& 238745.844 & 488.616 & \textcolor{red}{105.982} & 1776.269 
& 453917.428 & 673.734 & 129.605 & 3713.915 \\
ChatGPT-4o         & Zeroshot     & 
  209928.969 & 458.180 &  53.646 &  477.490 
& 262043.221 & 511.902 &  64.119 &  720.230 
& 692109.131 & 831.931 &  98.266 & 1003.000 \\
ChatGPT-4o         & Zeroshot+CoT & 
  304682.516 & 551.909 &  65.824 &  693.560 
& 254872.335 & 504.849 &  71.559 &  862.500 
& 700590.031 & 837.013 & 106.188 & 1082.000 \\
ChatGPT-4o         & Fewshot      & 
  212138.532 & 460.585 &  55.039 &  494.910 
& 270020.531 & 519.635 &  66.325 &  709.990 
& 692186.880 & 831.978 &  98.825 &  876.550 \\
ChatGPT-4o         & Fewshot+CoT  & 
  187493.536 & 433.005 &  53.400 &  573.980 
& 270299.315 & 519.902 &  68.360 &  909.600 
& 717158.980 & 846.852 & 102.963 &  935.810 \\
ChatGPT-4o         & Ablation     & 
  204751.320 & 452.495 &  52.713 &  458.360 
& 227027.901 & 476.474 &  69.114 &  853.210 
& 572783.332 & 756.824 &  98.364 &  888.440 \\
Gemini 1.5 Pro     & Zeroshot     & 
  304885.954 & 552.165 &  64.805 &  621.060 
& 429428.203 & 655.308 &  82.361 &  872.060 
& 1137572.307 & 1066.570 & \textcolor{red}{134.078} & 1168.070 \\
Gemini 1.5 Pro     & Zeroshot+CoT & 
  143821.533 & 379.238 &  \textcolor{blue}{47.026} &  477.670 
& 226087.288 & 475.486 &  63.860 &  811.710 
& 164604.723 & 405.715 &  \textcolor{blue}{60.745} &  873.390 \\
Gemini 1.5 Pro     & Fewshot      & 
  212138.532 & 460.585 &  55.039 &  494.910 
& 199224.822 & 446.346 &  66.204 &  768.430 
& 499077.529 & 706.454 &  91.128 &  862.460 \\
Gemini 1.5 Pro     & Fewshot+CoT  & 
  187493.536 & 433.005 &  53.400 &  573.980 
&  93385.020 & 305.590 &  \textcolor{blue}{56.793} & 1381.930 
& 319180.177 & 564.960 &  73.852 & 1054.980 \\
Gemini 1.5 Pro     & Ablation     & 
  204751.320 & 452.495 &  52.713 &  458.360 
& 227027.901 & 476.474 &  69.114 &  853.210 
& 572783.332 & 756.824 &  98.364 &  888.440 \\
\bottomrule
\end{tabular}
}
\caption{Baselines and LLMs' performance metrics across three subsets \textcolor{red}{Red: high value}, \textcolor{blue}{Blue: low value.}}
\label{tab:metrics_three_runs}
\end{table}



\begin{table}[ht]
\centering
\resizebox{1\textwidth}{!}{%
\begin{tabular}{l l *{8}{S}}
\toprule
Dataset & Prompt 
  & {EFR} & {AVGEFR} & {TTR} & {AVGTTR} & {DCR} & {AVGDCR} & {FVR} & {AVGFVR} \\
\midrule
15\_24\_100    & zeroshot     & 
  0.8811610077 & \textcolor{red}{0.874} & 0.006845564074 & 0.003947192219
  & 0.05832420591 & 0.09438143646 & 0 & 0.0002281021898 \\
15\_24\_50\textasciitilde   &              & 
  0.8817086528 & {}           & 0.004381161008 & {}
  & 0.09036144578 & {}            & 0 & {} \\
15\_24\_30\textasciitilde   &              & 
  0.8795620438 & {}           & 0.004562043796 & {}
  & 0.1332116788  & {}           & 0.0009124087591 & {} \\
15\_24\_10\textasciitilde   &              & 
  0.8524590164 & {}           & 0 & {}
  & 0.0956284153  & {}           & 0 & {} \\
\midrule
15\_24\_100    & zeroshot cot & 
  0.5506571742 & 0.5384918018 & 0.01040525739  & 0.004516575991
  & 0.06790799562 & 0.09441339935 & 0.001916757941 & 0.001254936801 \\
15\_24\_50\textasciitilde   &              & 
  0.5421686747 & {}           & 0.004928806134 & {}
  & 0.0991237678  & {}           & 0.002190580504 & {} \\
15\_24\_30\textasciitilde   &              & 
  0.5666058394 & {}           & 0 & {}
  & 0.1313868613  & {}           & 0.0009124087591 & {} \\
15\_24\_10\textasciitilde   &              & 
  0.4945355191 & {}           & 0.002732240437 & {}
  & 0.07923497268 & {}           & 0 & {} \\
\midrule
15\_24\_100    & few shot     & 
  0.4709748083 & 0.5112077674 & 0.09967141292 & 0.05757047985
  & 0.04518072289 & 0.08940042781 & 0.000273822563 & 0.0007984825833 \\
15\_24\_50\textasciitilde   &              & 
  0.5164293538 & {}           & 0.05038335159 & {}
  & 0.07995618839 & {}           & 0.001095290252 & {} \\
15\_24\_30\textasciitilde   &              & 
  0.5273722628 & {}           & 0.0447080292  & {}
  & 0.1231751825  & {}           & 0.001824817518 & {} \\
15\_24\_10\textasciitilde   &              & 
  0.5300546448 & {}           & 0.03551912568 & {}
  & 0.1092896175  & {}           & 0 & {} \\
\midrule
15\_24\_100    & few shot cot & 
  0.6004928806 & 0.5662320871 & 0.03258488499 & 0.01895772086
  & 0.05257393209 & 0.08377954282 & 0.00273822563 & 0.0006845564074 \\
15\_24\_50\textasciitilde   &              & 
  0.5805038335 & {}           & 0.02135815991 & {}
  & 0.08105147864 & {}           & 0 & {} \\
15\_24\_30\textasciitilde   &              & 
  0.5784671533 & {}           & 0.01642335766 & {}
  & 0.1195255474  & {}           & 0 & {} \\
15\_24\_10\textasciitilde   &              & 
  0.5054644809 & {}           & 0.005464480874 & {}
  & 0.08196721311 & {}           & 0 & {} \\
\midrule
15\_24\_100    & PURE zero    & 
  0.9529025192 & \textcolor{red}{0.950} & 0.007119386637 & 0.004607517061
  & 0.04682365827 & 0.0756363169  & 0 & 0.002965328467 \\
15\_24\_50\textasciitilde   &              & 
  0.9594742607 & {}           & 0.004928806134 & {}
  & 0.07338444688 & {}           & 0 & {} \\
15\_24\_30\textasciitilde   &              & 
  0.9397810219 & {}           & 0.003649635036 & {}
  & 0.1031021898  & {}           & 0.01186131387 & {} \\
15\_24\_10\textasciitilde   &              & 
  0.9480874317 & {}           & 0.002732240437 & {}
  & 0.07923497268 & {}           & 0 & {} \\
\bottomrule
\end{tabular}
}
\caption{ChatGPT-4o hallucination detection metrics; Echoing failure rate = EFR, Avarage echoing rate = AvgEFR; Trival transformation rate = TTR, Avarage trival transformation rate = AvgTTRDegenerate copy rage = DCR, Avarage degenerate copy rage = AvgDCR; Format violation rate = FVR, Avarage format violation rate = AvgFVR; }
\end{table}

\begin{table}[ht]
\centering
\resizebox{1\textwidth}{!}{%
\begin{tabular}{l l *{8}{S}}
\toprule
Dataset & Prompt 
  & {EFR} & {AVGEFR} & {TTR} & {AVGTTR} & {DCR} & {AVGDCR} & {FVR} & {AVGFVR} \\
\midrule
15\_24\_100    & zeroshot     & 
  0.1464950712  & 0.1733378356 & 0.01095290252 & 0.005337793842
  & 0.1311610077 & 0.1686830991 & 0 & 0.0110898138 \\
15\_24\_50\textasciitilde   &              & 
  0.1785323111  & {}           & 0.004928806134 & {}
  & 0.1642935378 & {}           & 0.0443592552 & {} \\
15\_24\_30\textasciitilde   &              & 
  0.2098540146  & {}           & 0.002737226277 & {}
  & 0.223540146  & {}           & 0 & {} \\
15\_24\_10\textasciitilde   &              & 
  0.1584699454  & {}           & 0.002732240437 & {}
  & 0.1557377049 & {}           & 0 & {} \\
\midrule
15\_24\_100    & zeroshot cot & 
  0.2234392114  & 0.192485675  & 0.003285870756 & 0.002099139757
  & 0.0517524644 & 0.06401357385 & 0.3012048193 & \textcolor{red}{0.364} \\
15\_24\_50\textasciitilde   &              & 
  0.1982475356  & {}           & 0.003285870756 & {}
  & 0.06024096386 & {}           & 0.3669222344 & {} \\
15\_24\_30\textasciitilde   &              & 
  0.1925182482  & {}           & 0.001824817518 & {}
  & 0.08941605839 & {}           & 0.3859489051 & {} \\
15\_24\_10\textasciitilde   &              & 
  0.1557377049  & {}           & 0 & {}
  & 0.05464480874 & {}           & 0.4016393443 & {} \\
\midrule
15\_24\_100    & few shot     & 
  0.01204819277 & 0.01190678986 & 0.01040525739 & 0.004746174479
  & 0.04600219058 & 0.07789535305 & 0.009583789704 & 0.002395947426 \\
15\_24\_50\textasciitilde   &              & 
  0.01369112815 & {}           & 0.007667031763 & {}
  & 0.08324205915 & {}           & 0 & {} \\
15\_24\_30\textasciitilde   &              & 
  0.01642335766 & {}           & 0.0009124087591 & {}
  & 0.1031021898  & {}           & 0 & {} \\
15\_24\_10\textasciitilde   &              & 
  0.005464480874& {}           & 0 & {}
  & 0.07923497268 & {}           & 0 & {} \\
\midrule
15\_24\_100    & few shot cot & 
  0.01341730559 & 0.05733585729 & 0.00547645126 & 0.003647891631
  & 0.02765607886 & 0.03783950459 & 0.3658269441 & \textcolor{red}{0.454} \\
15\_24\_50\textasciitilde   &              & 
  0.1922234392  & {}           & 0.00273822563 & {}
  & 0.06626506024 & {}           & 0.3674698795 & {} \\
15\_24\_30\textasciitilde   &              & 
  0.01277372263 & {}           & 0.0009124087591 & {}
  & 0.03284671533 & {}           & 0.5155109489 & {} \\
15\_24\_10\textasciitilde   &              & 
  0.01092896175 & {}           & 0.005464480874 & {}
  & 0.02459016393 & {}           & 0.5683060109 & {} \\
\bottomrule
\end{tabular}
}
\caption{Gemini 1.5 Pro hallucination detection metrics}
\end{table}

Australia’s electricity market has not always been stable\cite{qu2018volatilityMarket}. From 2015 to 2024, it has shown increasing signs of volatility in NSW's electricity market\cite{begin2025StochasticElectricityPrice}. Electricity prices can sometimes exhibit extreme spikes or drops, posing significant challenges for LLMs to forecast such events accurately before they occur. From our observation of the electricity price curve over the ten-year period, we found that such extreme events became more frequent as the data approached the end of 2024. Therefore, we divided the ten-year dataset into multiple groups of different sizes to better analyze model performance under varying data conditions. We evaluated the performance of both baseline models and LLMs on the full dataset, as well as on the most recent 50 percent, 30 percent and 10 percent of the whole dataset. By doing this we will be able to analyse LLM's performance on different period of the whole dataset.

\paragraph{Baselines. }Since the test sets for the machine learning models do not cover the full dataset, we compared their performance with that of the LLMs using the same subsets (recent 50\%, 30\%, 10\%). The results show that among the traditional baselines, ARIMA consistently achieves the lowest MAE across all three dataset portions. Linear Regression records the lowest MAPE in each subset, while XGBoost performs the worst across all evaluation metrics. A common pattern observed is that as the size of the dataset subset decreases, the error scores of the baseline models tend to increase. For both ARIMA and Linear Regression, their MAE scores increase by approximately 30 points when moving from the 50\% subset to the 10\% subset, but XGBoost shows a comparatively more stable increase in MAE, with a rise of around 16 points.

\paragraph{ChatGPT-4o and Gemini 1.5Pro.} Despite the incorporation of auxiliary information—including contemporaneous price trends, news sentiment, and temperature data—and the application of four distinct prompt-engineering paradigms, the forecasting performance of ChatGPT-4o and Gemini 1.5 Pro remains inconsistent and, in several instances, inferior to classical statistical baselines. On the 50\% data subset, for example, ARIMA outperforms ChatGPT-4o under both zero-shot-plus-CoT and few-shot configurations, and likewise surpasses Gemini 1.5 Pro under zero-shot and few-shot settings. The ablation experiment, designed to disentangle the marginal utility of background knowledge, further underscores this limitation: none of the engineered prompts elevates ChatGPT-4o beyond its ablation counterpart, and only Gemini’s zero-shot-plus-CoT variant marginally exceeds its own ablation baseline. Even so, the best individual error metrics still originate from the LLMs: on the 50\% subset, Gemini 1.5 Pro paired with zero-shot-plus-CoT attains the lowest MSE, RMSE, and MAPE, while its MAE (477.67) trails the overall minimum—ChatGPT-4o’s zero-shot MAE of 477.49—by a negligible 0.18 points; similar isolated gains appear on the 30\% and 10\% subsets. Nevertheless, traditional accuracy scores alone prove insufficient, as hallucination analyses reveal pervasive reliability concerns. Under an ablation prompt containing only historical prices, ChatGPT-4o exhibits an average echoing-error rate of 95\% across all dataset partitions, falling only to 53 – 56\% with advanced prompting, while trivial‐transformation and degenerate-copy errors peak at 5.7\% and 9.4\%, respectively; format violations, though rare (<0.1\%), remain non-negligible. Gemini 1.5 Pro demonstrates superior echoing control—dropping from 32\% in ablation to as low as 1.19\% under few-shot prompts—but at the expense of formatting robustness: CoT prompts provoke format-violation rates of 36.39\% (zero-shot) and 45.54\% (few-shot). Although trivial transformation stays below 1\% and degenerate copying declines to 3.78\% in few-shot-CoT, the modest accuracy improvements may partly reflect the censoring of invalid outputs rather than genuine predictive gains. Collectively, these findings indicate that—even when enriched with exogenous context and sophisticated prompting strategies—current LLM-based forecasters exhibit volatile accuracy and persistent hallucination behaviour, rendering their practical reliability uncertain relative to established statistical models.

\section{Conclusion \& Discussion}\label{conclusion}
We introduce \textsc{NSW-EPNews}, the first benchmark that fuses half-hourly spot prices with curated news summaries and temperature readings, enabling head-to-head evaluation of classical forecasters and state-of-the-art LLMs on a genuinely multimodal electricity-price task. Extensive tests across four prompt regimes and multiple temporal splits reveal that today’s leading LLMs neither surpass ARIMA nor a linear baseline and remain susceptible to systemic hallucinations—echoing inputs, offset shifts, degenerate repeats, and format failures. These findings expose a substantial gap between present LLM capabilities and the reliability required for high-stakes energy forecasting, and they position \textsc{NSW-EPNews} as a principled test-bed for future work on prompt design, retrieval grounding, and hallucination mitigation in time-series applications.

\paragraph{Limitations. }NSW\textendash EPNews, while providing the first news-augmented benchmark for electricity-price forecasting, is constrained in four respects: (i) all news summaries are produced by GPT-4o, so the impact of alternative LLM summarisers on data fidelity and downstream accuracy remains unknown; (ii) the prompt design space is restricted to four variants (zero/few-shot with or without chain-of-thought), leaving more advanced schemes—such as retrieval-augmented or self-consistency prompts—unexplored; (iii) our hallucination analysis tracks only four surface-level failure modes (echoing, offset shifts, degenerate copies, format violations), omitting deeper semantic or causal errors; and (iv) empirical evaluation covers two proprietary LLMs, leaving the behaviour of open-source and domain-specialised models to future work.

\paragraph{Future work. }  Future enhancements will focus on: (i) refining prompt engineering by exploring more sophisticated structures, dynamic guidance, and adaptive prompting strategies; (ii) assessing the suitability of open-source large models, domain-specific lightweight models and AI agent for electricity-price forecasting; (iii) expanding data coverage by incorporating policy bulletins, social-media sentiment, and other heterogeneous textual signals; and (iv) enhancing multimodal fusion—e.g., retrieval-augmented generation (RAG) and graph neural networks to model event relationships—to further boost predictive accuracy.

\paragraph{Key Findings.}
Although LLMs excel at natural-language understanding, our experiments show they remain brittle forecasters. Long, multimodal prompts frequently trigger hallucinations—echoed prices, constant offsets, or repeated values—and, under chain-of-thought prompting, some models even announce an “ARIMA step,” betraying reliance on canned heuristics rather than genuine reasoning. Injecting news summaries, temperature data, and refined prompts mitigates these pathologies only marginally: on every split the two state-of-the-art LLMs record higher MAE/RMSE than both classical baselines and an ablation that uses historical prices alone, while vectorised news affords traditional models little extra accuracy. Error magnitudes rise sharply as the test window narrows to recent years, peaking on the final 10 \% of data where price spikes are most frequent; conversely, evaluation over the full 2015–2024 horizon dilutes such outliers and yields RMSE values in the low thirties. Taken together, the results underline two risks: existing ML baselines still struggle with extreme volatility, and current LLMs—despite sophisticated prompt engineering—cannot yet deliver reliable, news-aware electricity-price forecasts in high-stakes settings.

\newpage
\bibliography{ref}
\newpage

\appendix
\section{Prompt settings}\label{app:promtsettings}
\paragraph{Zero-shot.}
The zero-shot style prompt includes key contextual information such as one day's worth of historical electricity price data (48 half-hourly points), the summarized news published on that day, and additional background details including public holiday indicators, maximum and minimum temperatures. The expected output format of the LLM was also explicitly specified in the prompt to ensure consistency and ease of evaluation. This prompt format does not include any examples or explicit reasoning instructions, relying instead on the LLM’s generalization ability to produce accurate forecasts for next day's electricity price based solely on the given context.

\paragraph{Few-shot.}
Building on the zero-shot style prompt, two example question–answer pairs were added to create a few-shot prompt format. This approach is intended to guide the LLM toward generating more consistent and structured responses, while also reducing the likelihood of hallucinations or false responses by providing clear demonstrations of the expected input–output relationship.

\paragraph{Zero-shot + CoT.}
This approach still follows the zero-shot style but integrated with chain of thought prompting. Chain-of-thought prompting is one of the most commonly used techniques for guiding LLMs toward better reasoning and prediction performance. By providing step-by-step instructions on how to reason about future electricity prices—such as considering historical trends, interpreting relevant news, and accounting for contextual factors—this prompt format encourages the LLM to produce more logical and transparent predictions.

\paragraph{Few-shot + CoT.}
Both few-shot and chain-of-thought prompting were integrated into the base prompt. Specifically, two illustrative examples containing sample reasoning steps were added to demonstrate how to analyze historical price trends, interpret news content, and arrive at a forecast. This combined approach aims to maximize the reliability of the LLM’s output by reducing hallucinations and enhancing forecasting accuracy.

\paragraph{Ablation Study.}
Before applying the four prompt engineering techniques, we first constructed a zero-shot style prompt that excludes any news or temperature information. This prompt allows the LLMs to perform forecasting based solely on the price history, without any additional contextual input or prompt engineering strategies. This ablation study is designed to highlight the impact of incorporating news and temperature information on the forecasting performance of LLMs

\begin{figure}[htbp]
  \centering
  \includegraphics[width=\linewidth,page=1]{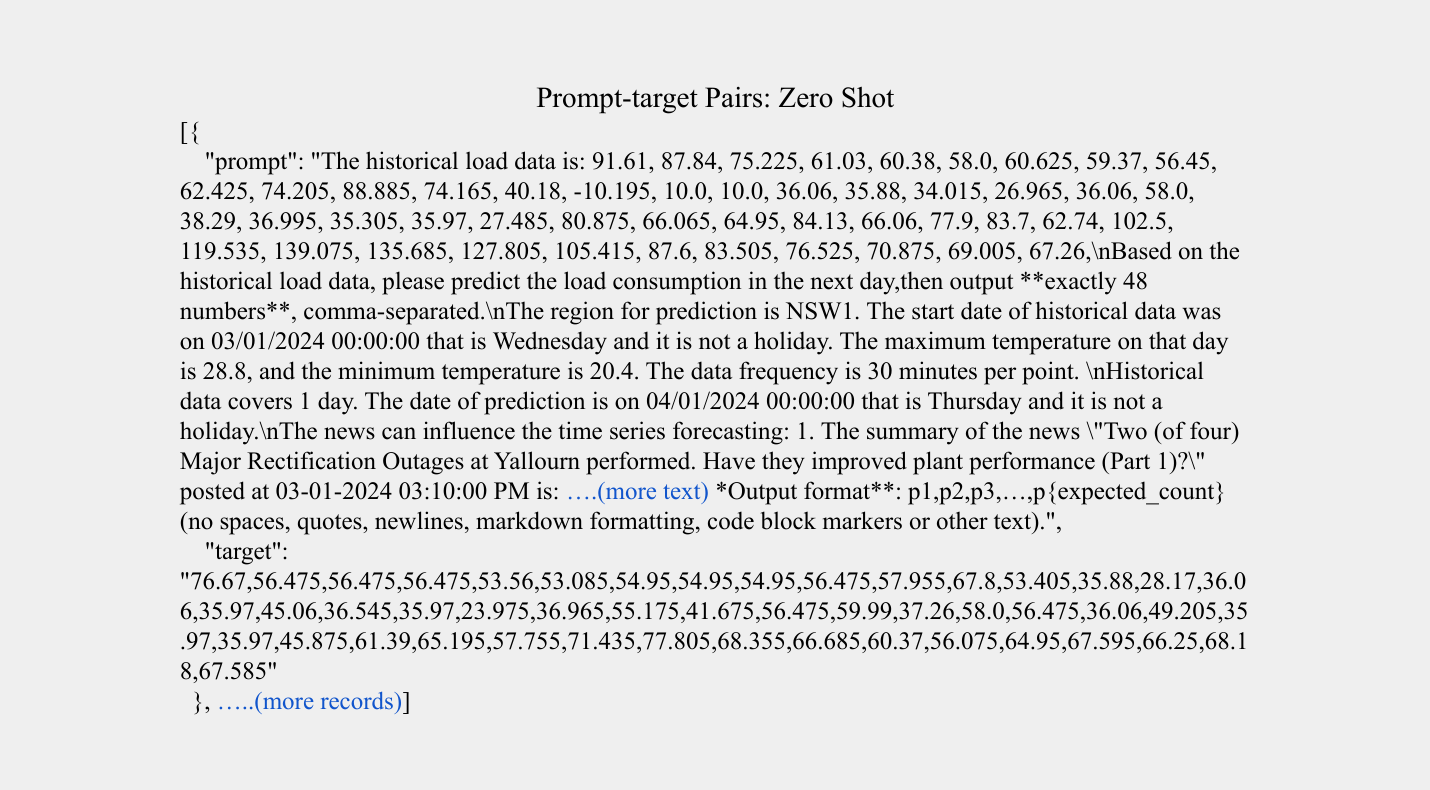}
  \caption{Zero-shot}
  \label{fig:example}
\end{figure}
\begin{figure}[htbp]
  \centering
  \includegraphics[width=\linewidth,page=1]{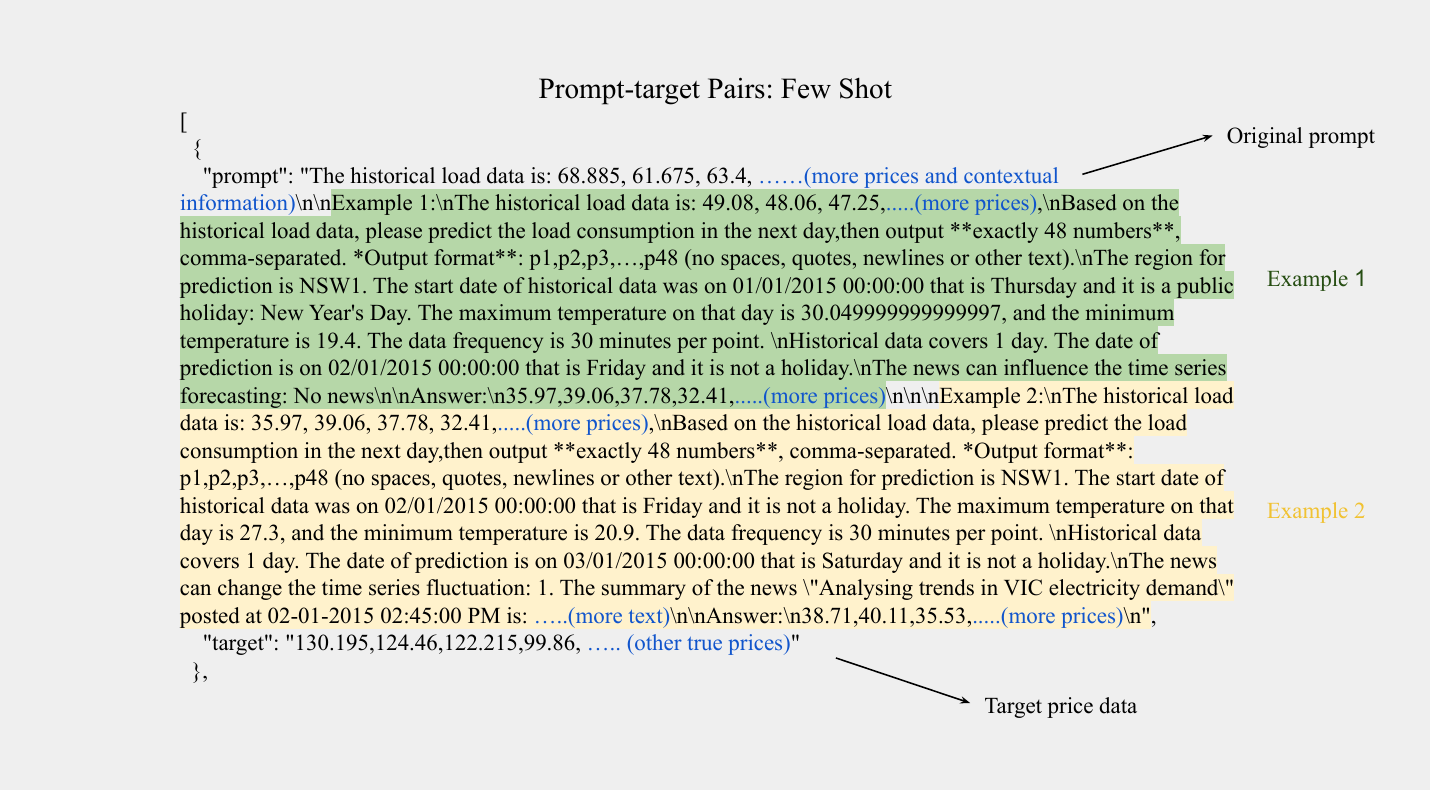}
  \caption{Few-shot}
  \label{fig:example}
\end{figure}
\begin{figure}[htbp]
  \centering
  \includegraphics[width=\linewidth,page=1]{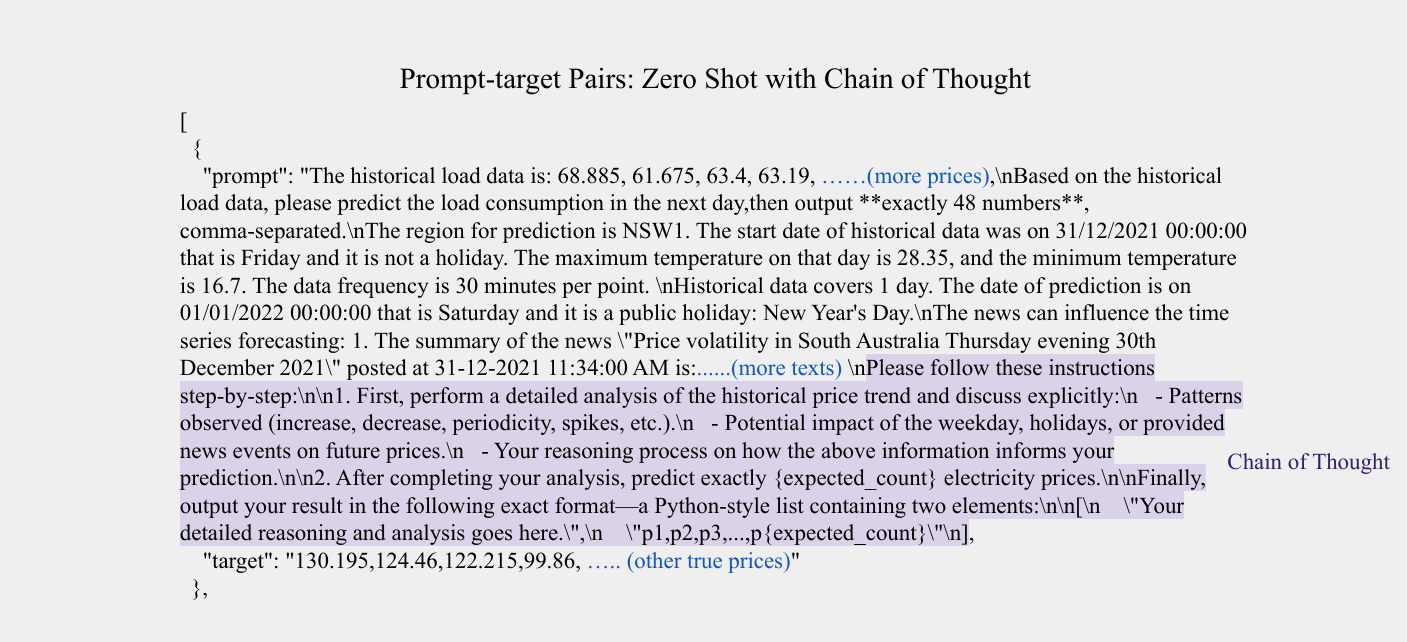}
  \caption{Zero-shot + CoT}
  \label{fig:example}
\end{figure}
\begin{figure}[htbp]
  \centering
  \includegraphics[width=\linewidth,page=1]{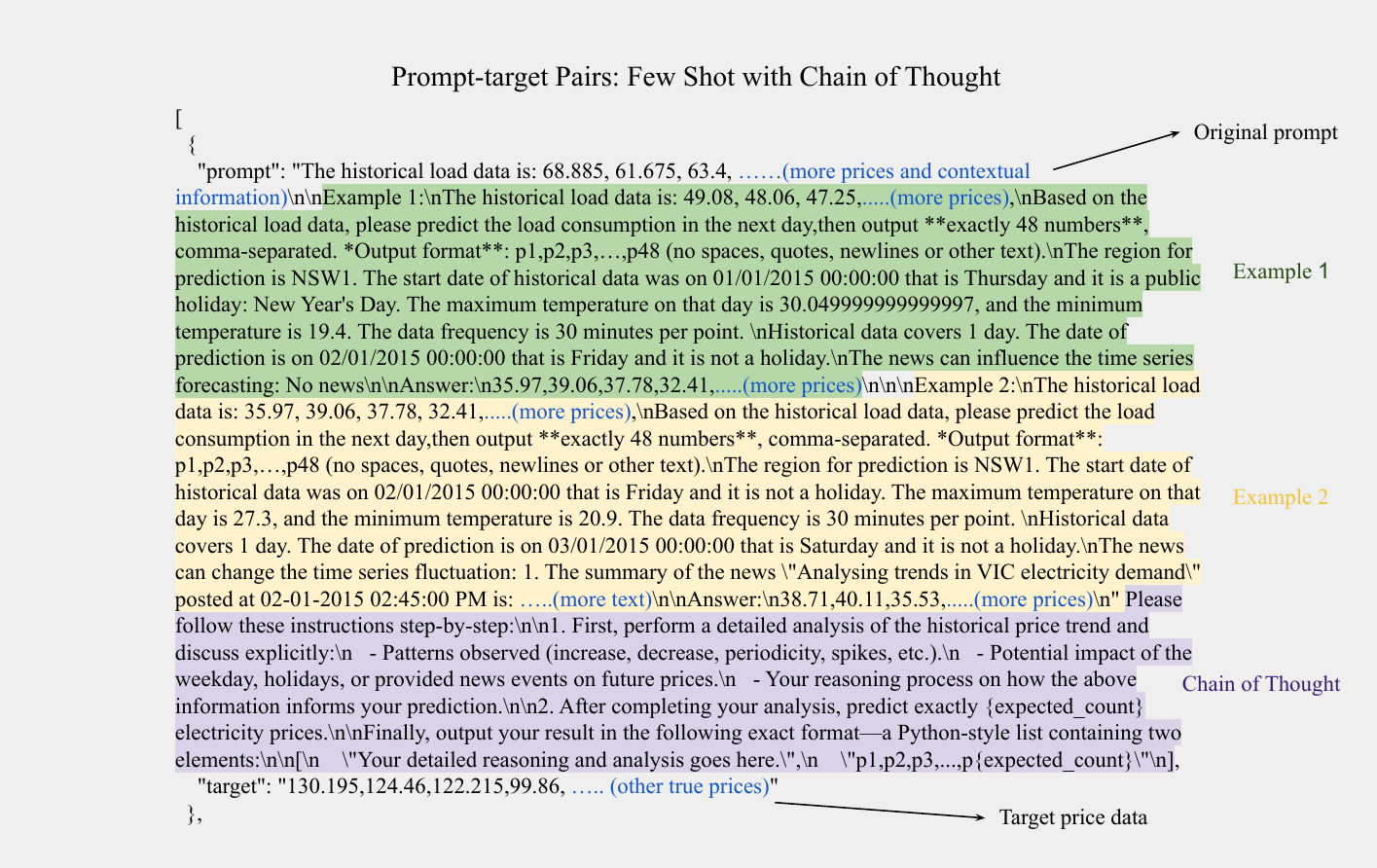}
  \caption{Few-shot + CoT}
  \label{fig:example}
\end{figure}

\section{Hallucination and error examples}\label{app:hallucination}
The hallucination and error outputed by LLMs in the experiment are defined by us as below:
\begin{itemize}
    \item \textbf{Echoing Failure}\hspace{1em}Although our prompts explicitly instruct the model to generate forecasts based on an analysis of historical electricity prices and contextual information, the model occasionally echoes segments of the historical data verbatim and presents them as predictions. This behaviour indicates that it is not truly reasoning about future prices, instead, the responses constitute spurious outputs. For each iteration, if ten or more of the LLM's predicted prices are exactly the same as the historical price data, the iteration is considered to have triggered an echoing failure.

    \item \textbf{Trivial Transformation}\hspace{1em}Similarly, even when the model avoids repeating the input verbatim,, it sometimes produces forecasts by merely applying a uniform offset to the historical prices. It reveals superficial pattern matching rather than genuine reasoning. If at least 20 of the LLM's predicted values can be obtained by simply adding or subtracting a constant positive number from the historical data, that generation is identified as a trivial transformation case.

    \item \textbf{Degenerate Copying}\hspace{1em}In other instances, the model produces the same value repeatedly sometimes for the entire forecast horizon. Although the official electricity-price records occasionally contain stretches of unchanged prices, such flat periods are relatively uncommon. It remains important to record these cases in which the model repeats a single value across its forecasts. We identify a degenerate copying case when the LLM's predicted values are repeating one number for over five times.

    \item \textbf{Format Violation}\hspace{1em}LLMs are instructed to output their predicted electricity prices and sometimes along with reasoning steps in specific formats predefined by us. However, in some cases, the models fail to follow the required format, resulting in parsing errors when trying to extract their predicted prices. We record those failures for each LLM when processing prompts. This metric reflects the consistency and reliability of the model’s output format, which is a crucial factor in real-world applications where structured post-processing is necessary. If the LLM's answer violates the output format, there will be a parsing error.
\end{itemize}

\begin{figure}[htbp]
  \centering
  \includegraphics[width=\linewidth,page=1]{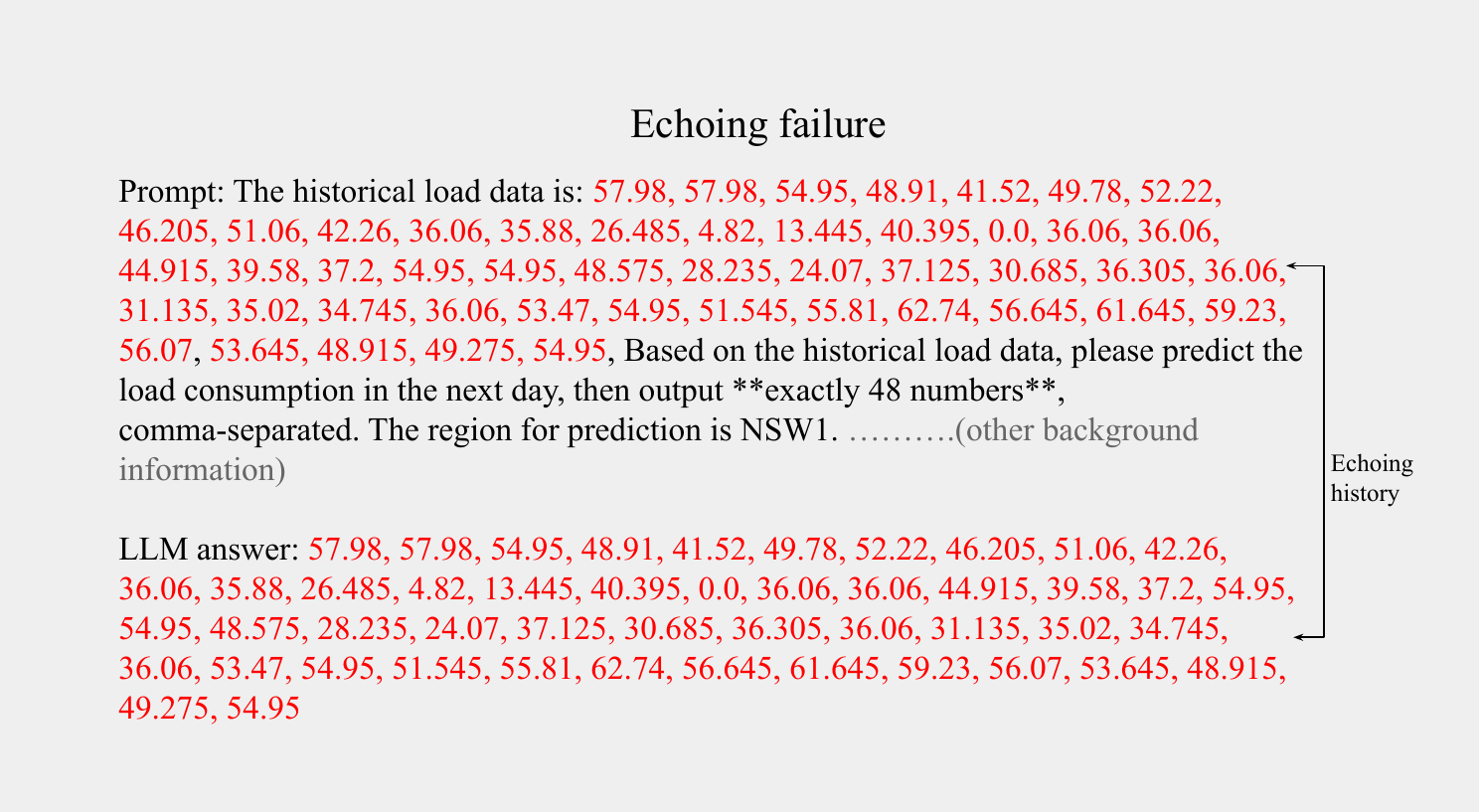}
  \caption{Echoing Failure}
  \label{fig:example}
\end{figure}
\begin{figure}[htbp]
  \centering
  \includegraphics[width=\linewidth,page=1]{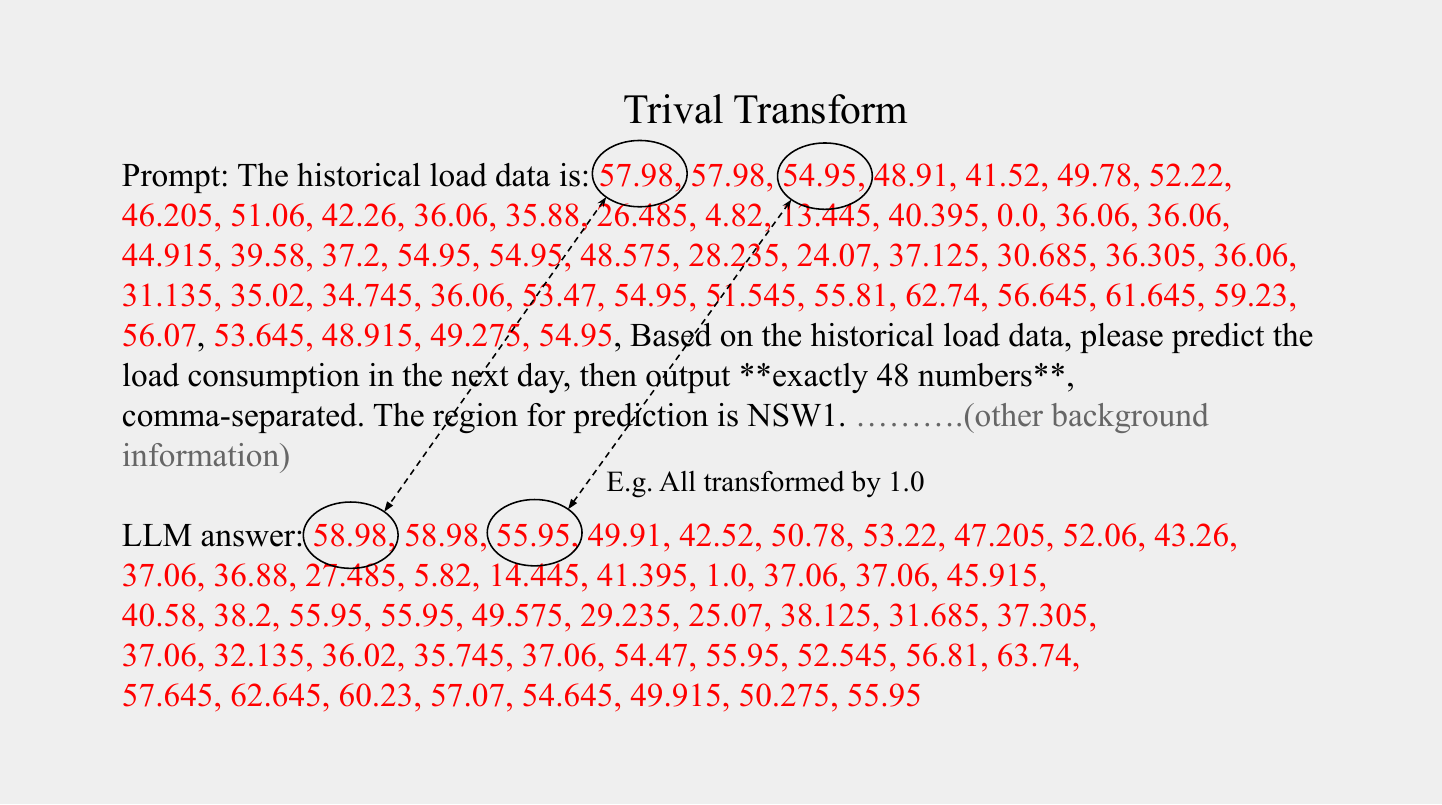}
  \caption{Trival Transformation}
  \label{fig:example}
\end{figure}
\begin{figure}[htbp]
  \centering
  \includegraphics[width=\linewidth,page=1]{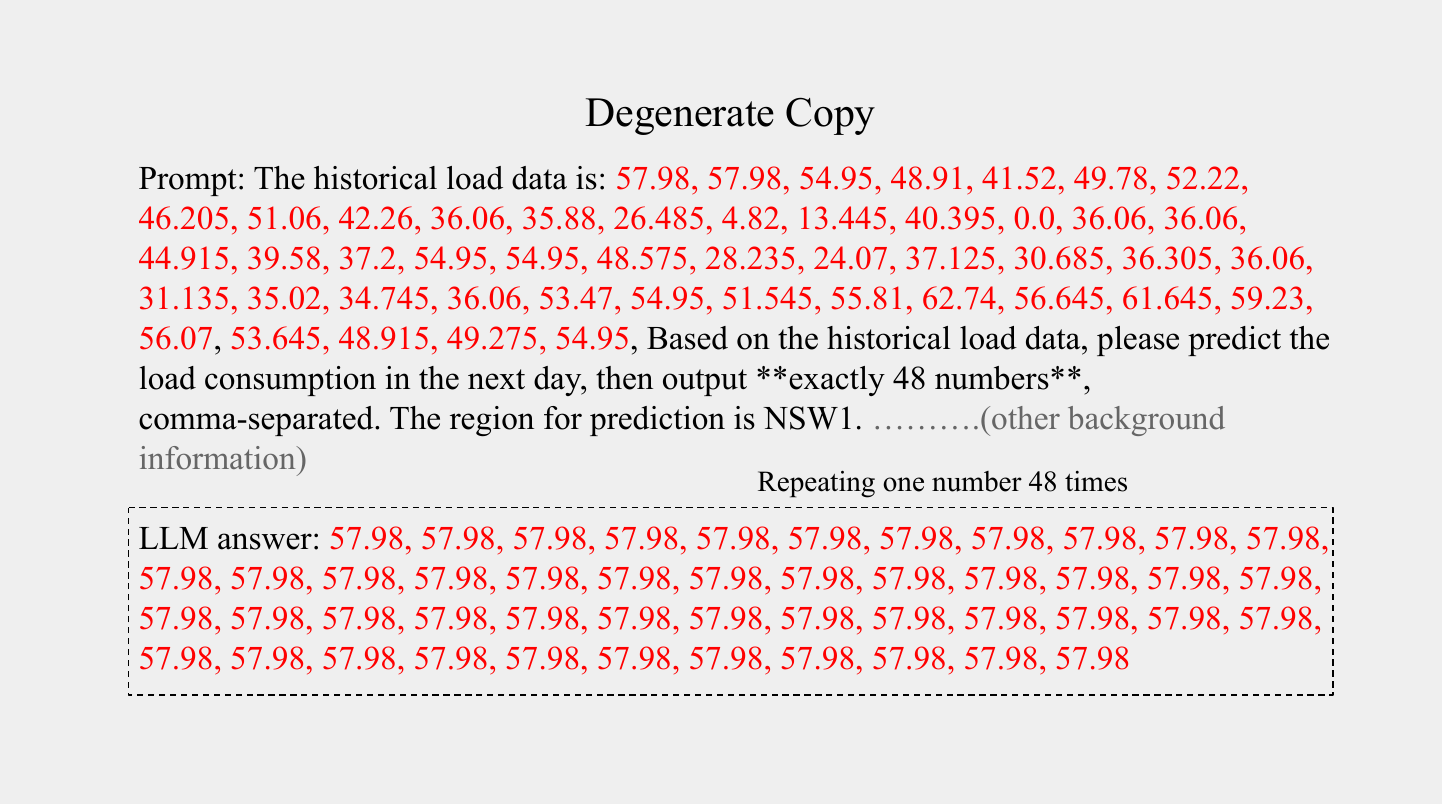}
  \caption{Degenerate Copy}
  \label{fig:example}
\end{figure}
\begin{figure}[htbp]
  \centering
  \includegraphics[width=\linewidth,page=1]{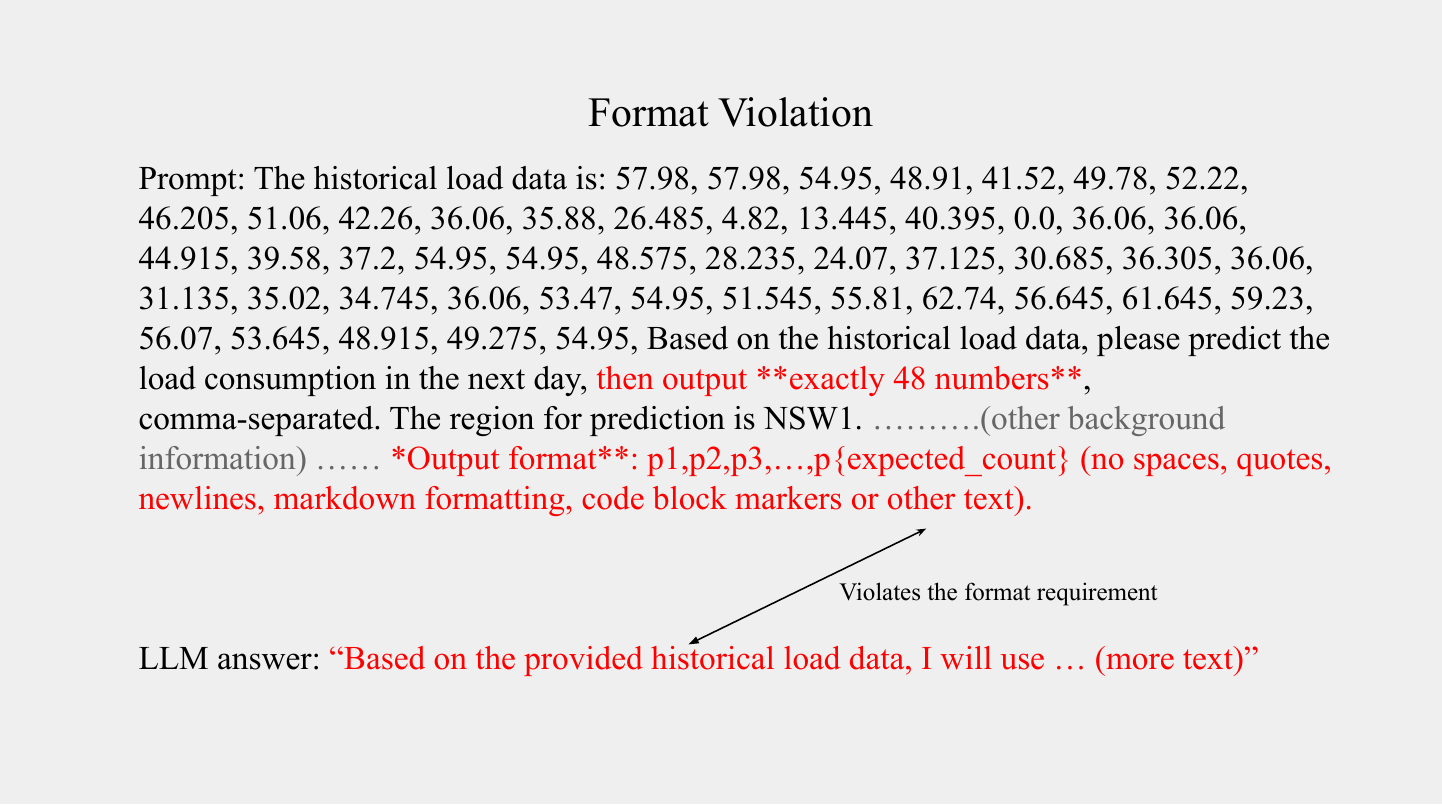}
  \caption{Format Violation}
  \label{fig:example}
\end{figure}

\section{Experiment Settings}\label{app:experimentsettings}
\subsection{Baseline models}
For baseline models, we selected the statistical model ARIMA and two machine learning methods, LR and XGBoost. These models represent traditional approaches commonly used in time-series forecasting and serve as reference points for comparison with modern state-of-the-art LLMs. Machine learning based methods utilized the raw dataset and are evaluated on the same subsets as ARIMA does. We use Linear Regression as a transparent linear baseline and XGBoost as a nonlinear tree ensemble. Each model ingests a feature vector that combines the last ten prices, a fifty-dimensional TF-IDF representation of the accompanying news, and the day’s minimum and maximum temperatures. When the most recent 50\% (from 2020 to the end of 2024), 30\% (from 2022 to the end of 2024), and 10\% (the year 2024) of the full dataset were used as test sets, the remaining portions of the dataset served as the corresponding training sets.

\begin{itemize}
    \item \textbf{ARIMA}: We used the raw dataset to evaluate ARIMA as a classical, fully-numerical baseline. The model was applied to the full dataset as well as its most recent 50\%, 30\%, and 10\% subsets (spanning 2020–2024, 2022–2024, and 2024 respectively). For each test record, an ARIMA(1, 1, 1) model was fit to the half-hourly historical price series and used to forecast the next 48 half-hour steps. ARIMA relies solely on autoregressive dynamics in the price series, ignoring weather and news covariates. The forecasts were evaluated using standard error metrics such as MAE and RMSE, providing a baseline for comparing multimodal and LLM-based approaches.
    
    \item \textbf{Linear Regression}: This transparent linear model was tested on the same subsets as ARIMA. It receives a feature vector built from: (1) the last 10 half-hour prices, (2) a 50-dimensional TF-IDF representation of the day’s news, and (3) the day’s minimum and maximum temperatures (with missing temperature values filled with the global mean). The dataset was windowed to create training samples, and each model was trained once on the corresponding training set. At test time, a rolling forecast was performed: the model predicted one step at a time for 48 steps, continuously updating its input window. Evaluation was based on MAE, MSE, RMSE, and MAPE.
    
    \item \textbf{XGBoost}: As a nonlinear tree-ensemble method, XGBoost was configured with 100 trees, a maximum depth of 5, and a learning rate of 0.1. It used the same input features as Linear Regression (price window + TF-IDF news + temperatures). Like Linear Regression, it was trained on windowed samples and evaluated using rolling forecasts on the same data splits. This model aims to capture non-linear dependencies that may be missed by simpler linear models or autoregressive baselines.
\end{itemize}

\subsection{Large language models}
We selected two state-of-the-art closed-source LLMs, ChatGPT-4o\cite{openai2024gpt4o} and Gemini 1.5 Pro\cite{reid2024gemini}. They will be evaluated using the constructed prompt–target pairs. Each LLM was tested with zero-shot, few-shot, zero-shot+CoT, few-shot+CoT prompt-engineering strategies, and under each strategy, the model was evaluated on the full dataset as well as the most recent 50\%, 30\%, and 10\% subsets. During evaluation, prompts were fed to the models via iterative API calls. The generated responses will be compared with the target values to calculate error scores. In addition, the generated responses will also be compared against the provided historical price data to identify instances of hallucinations and erroneous outputs. This process includes checking whether the generated responses follow the required output format. If the format is correct, we further examine whether the responses simply echo the historical prices, apply a fixed offset to them, or consist of a repeated single value throughout the forecast sequence. Then LLMs' predicted price sequence will be extracted and scored against the true prices using MAE, MSE, RMSE and MAPE. 

\section{News article example from WattClarity}\label{app:News article}
\begin{itemize}
    \item On Wednesday 27th November 2024 2:48 PM, a news titled with "NSW Prices spike early ahead of forecasts on afternoon of 27th November" posted: "A quick note to record that Energy price in NSW has already spiked to up near the market price cap of \$17500 as early as 14:30 NEM time, seen here in Forecast Convergence grid view. Pricing outcomes on high demand days like this are clearly driven by a number of complex factors, but increasingly we are finding value in the overall story being told by our Congestion Map prototype."\cite{kent2024WattNews1}
    \item On Friday 8th November 2024 7:03 PM, a news titled with "A much shorter run of evening volatility in QLD and NSW on Friday 8th November 2024" posted "
    It was more humid on the walk home from the Brisbane office late this afternoon … but at least there were not as many SMS pings as there were for price alerts on Thursday 7th November 2024, with the following being the capture of prices in any region above \$1,000/MWh...."\cite{mcardle2024WattNews2}
\end{itemize}

\section{Evaluation metrics}\label{app:Evaluationmetrics}
For both baseline models and LLMs, we compare their predicted prices with thte true prices one by one and calcualte the standard metrics.
Note that LLMs may occasionally produce slightly more or fewer predicted price values than expected. In such cases, any extra predictions are discarded. Missing values are initially filled with null placeholders and later excluded, along with their corresponding ground truth values, during the calculation of error metrics. We used four standard evaluation metrics: mean squared error (MSE), mean absolute error (MAE), root mean squared error (RMSE), and mean absolute percentage error (MAPE).

\begin{itemize}
  \item \textbf{Mean Squared Error (MSE)}: MSE is sensitive to large errors and thus emphasizes the impact of extreme values in electricity price forecasting. It is calculated by averaging the squares of the differences between predicted and actual prices.
  
  \item \textbf{Root Mean Squared Error (RMSE)}: RMSE is the square root of MSE. It inherits MSE’s sensitivity to large deviations, providing error in the same unit as the original price (e.g., AUD), which aids interpretability.
  
  \item \textbf{Mean Absolute Error (MAE)}: MAE reflects the average absolute difference between the predicted and actual prices. It provides a straightforward measure of the typical prediction error in Australian dollars and is less sensitive to outliers compared to MSE and RMSE.
  
  \item \textbf{Mean Absolute Percentage Error (MAPE)}: MAPE expresses the prediction error as a percentage of the actual price, offering an intuitive understanding of relative error. However, it can become unstable or undefined when the true prices are close to zero, which occasionally happens in the Australian electricity market. In practice, zero-value true prices are removed from the dataset when computing MAPE.
\end{itemize}

The formulas are shown below, where $y_i$ denotes the $i$-th true value, $\hat{y}_i$ denotes the corresponding predicted value, and $N$ is the total number of samples.


\[
\begin{alignedat}{2}
\mathrm{MAE}  &= \frac{1}{N}\sum_{i=1}^{N}\lvert \hat{y}_i - y_i \rvert
\qquad &
\mathrm{MSE}  &= \frac{1}{N}\sum_{i=1}^{N}(\hat{y}_i - y_i)^2 \\[6pt]
\mathrm{RMSE} &= \sqrt{\frac{1}{N}\sum_{i=1}^{N}(\hat{y}_i - y_i)^2}
\qquad &
\mathrm{MAPE} &= \frac{100\%}{N}\sum_{i=1}^{N}
                 \left|\frac{\hat{y}_i - y_i}{y_i}\right|
\end{alignedat}
\]

\section{News classifying prompt}\label{app:NewsclassifyPrompt}
The prompt that used by ChatGPT-4o to summarize news including following key points.

\begin{itemize}[leftmargin=*]
    \item \textbf{Role assignment.}  
    The prompt begins with \emph{“You are an expert in electricity market analysis”}.  
    This persona priming has two effects: (i) it biases the model toward domain-relevant vocabulary and causal reasoning (e.g.\ linking turbine failure to reserve scarcity), and (ii) it reduces generic, copy-editing style hallucinations by anchoring generation in an expert voice.
    
    \item \textbf{Classification criteria.}  
    We supply an explicit three-level taxonomy—\textsc{Level 1} catastrophic outages or fuel shortages, \textsc{Level 2} operational or price signals, \textsc{Level 3} policy or sentiment items.  
    Embedding the taxonomy inside the prompt forces the model to issue a categorical relevance label, enabling us to stratify the benchmark and later test whether forecasters place higher weight on \textsc{Level 1} events than on routine policy news.
    
    \item \textbf{Key attributes.}  
    The prompt enumerates ten named fields (\textit{timeframe of impact}, \textit{root cause}, \textit{affected region}, etc.).  
    These slots compel GPT-4o to transform unstructured prose into a fixed schema that is readily ingested by downstream models and analytic scripts.  
    When an attribute is absent in the article, the model must output \texttt{Unknown}, making missingness explicit and avoiding silent information leakage.
    
    \item \textbf{Summary rules.}  
    Finally, a rigid output template (length $\le 30{,}000$ characters, keyword–value pairs, date self-verification) standardises length and formatting.  
    This constraint not only simplifies post-processing but also limits free-form text that could introduce syntactic noise or hallucinated details.
\end{itemize}

\section{Experiment Results}\label{app:experiment results}
\textbf{Echoing failure rate = EFR, Avarage echoing rate = AvgEFR.}
The echoing failure rate on each dataset was calculated by dividing the number of times the LLM triggered an echoing failure while processing prompts by the total number of records in the prompt dataset. The average echoing failure rate of a prompt setting is calculated by summing the echoing failure rates across the four different dataset portions under that setting and then computing the mean value.

\textbf{Trival transformation rate = TTR, Avarage trival transformation rate = AvgTTR.} 
The trival transformation rate on each dataset was calculated by dividing the number of times the LLM triggered a trival transformation while processing prompts by the total number of records in the prompt dataset. The average trival transformation rate of a prompt setting is calculated by summing the trival transformation rates across the four different dataset portions under that setting and then computing the mean value.

\textbf{Degenerate copy rage = DCR, Avarage degenerate copy rage = AvgDCR.} 
The degenerate copy rate on each dataset was calculated by dividing the number of times the LLM triggered a degenerate copy while processing prompts by the total number of records in the prompt dataset. The average degenerate copy rate of a prompt setting is calculated by summing the degenerate copy rates across the four different dataset portions under that setting and then computing the mean value.

\textbf{Format violation rate = FVR, Avarage format violation rate = AvgFVR.} 

The format violation rate on each dataset was calculated by dividing the number of times the LLM triggered n format violation while processing prompts by the total number of records in the prompt dataset. The average format violation rate of a prompt setting is calculated by summing the format violation rates across the four different dataset portions under that setting and then computing the mean value.

In the ablation experiment, ChatGPT-4o uses only historical prices. It shows an average echoing failure rate of 95\%. When we apply prompt engineering, echoing rates fall but remain high. In our experiments with the four different prompt engineering techniques, ChatGPT-4o exhibits significant echoing issues. The zero-shot setting shows an average echoing failure rate of 87.37\%. Adding few-shot examples or chain-of-thought (CoT) reasoning reduces echoing considerably. However, the average echoing failure rate of ChatGPT-4o across the four dataset proportions remains relatively high, approximately 53\%-56\%. Forecasting accuracy measured by standard evaluation metrics deteriorates as the dataset size decreases. This decline is likely due to increased volatility and unseen data patterns in recent datasets. Echoing problems dominate other error types. Trivial transformations, such as fixed-value offsets, peak around 5.7\% under few-shot prompts. Degenerate copying occurs moderately, ranging from 8.3\% to 9.4\%. Some overlap with echoing cases may exist due to naturally occurring repeated values in electricity price data. Despite these issues, ChatGPT-4o maintains high output format stability. Format violation rates remain consistently below 0.1\%, even under CoT prompts. Despite reducing echoing, none of the prompt styles improve accuracy compared to the ablation setting. Their error scores remain higher than those of the ablation baseline.

In the ablation experiment, Gemini’s average echoing failure rate is about 32\%. Prompt engineering further reduces echoing but increases the average format violation rate. Gemini 1.5 Pro triggers fewer echoing failure rate compared to ChatGPT-4o. Yet, Gemini still performs poorly based on standard accuracy metrics. This suggests undetected or subtle hallucinations persist. Gemini's primary limitation is the high format-violation rate under CoT prompts. It averages 36.39\% for zero-shot CoT and 45.54\% for few-shot CoT across the four dataset proportions. Gemini has moderate echoing failure rates in zero-shot settings, ranging from 17.33\% to 19.24\%. Adding few-shot examples significantly reduces these rates, bringing them down to an average of 1.19\%-5.73\%. Trivial transformation rates remain negligible, below 1\%. Degenerate copying peaks at 16.86\% in zero-shot scenarios but declines below 10\% with CoT and few-shot prompts. The few-shot CoT scenario achieves the lowest degenerate copying rate at 3.78\%. Overall, Gemini exhibits fewer hallucination-related errors. However, its considerable formatting instability under complex prompting substantially limits its practical reliability. Accuracy metrics improve under all prompt styles except zero-shot, with modest decrease in MAE, MSE, RMSE, and MAPE over Gemini's ablation and ChatGPT-4o. However, it remains unclear whether the observed improvement in accuracy is a result of genuine model performance or simply due to the exclusion of invalid, improperly formatted LLM responses.

The ablation setting, which excludes news and temperature data, performs worse than ARIMA and Linear Regression on all four metrics: MAE, MSE, RMSE, and MAPE—and only outperforms XGBoost. Even after incorporating news and temperature information, the performance of LLMs under certain prompt settings still fails to surpass that of the baseline models ARIMA and LR. The result shows that adding news, temperature, and advanced prompts helps curb blatant hallucinations but does not guarantee better numeric forecasts for LLMs.

\newpage
\section*{NeurIPS Paper Checklist}

The checklist is designed to encourage best practices for responsible machine learning research, addressing issues of reproducibility, transparency, research ethics, and societal impact. Do not remove the checklist: {\bf The papers not including the checklist will be desk rejected.} The checklist should follow the references and follow the (optional) supplemental material.  The checklist does NOT count towards the page
limit. 

Please read the checklist guidelines carefully for information on how to answer these questions. For each question in the checklist:
\begin{itemize}
    \item You should answer \answerYes{}, \answerNo{}, or \answerNA{}.
    \item \answerNA{} means either that the question is Not Applicable for that particular paper or the relevant information is Not Available.
    \item Please provide a short (1–2 sentence) justification right after your answer (even for NA). 
\end{itemize}

{\bf The checklist answers are an integral part of your paper submission.} They are visible to the reviewers, area chairs, senior area chairs, and ethics reviewers. You will be asked to also include it (after eventual revisions) with the final version of your paper, and its final version will be published with the paper.

The reviewers of your paper will be asked to use the checklist as one of the factors in their evaluation. While "\answerYes{}" is generally preferable to "\answerNo{}", it is perfectly acceptable to answer "\answerNo{}" provided a proper justification is given (e.g., "error bars are not reported because it would be too computationally expensive" or "we were unable to find the license for the dataset we used"). In general, answering "\answerNo{}" or "\answerNA{}" is not grounds for rejection. While the questions are phrased in a binary way, we acknowledge that the true answer is often more nuanced, so please just use your best judgment and write a justification to elaborate. All supporting evidence can appear either in the main paper or the supplemental material, provided in appendix. If you answer \answerYes{} to a question, in the justification please point to the section(s) where related material for the question can be found.

IMPORTANT, please:
\begin{itemize}
    \item {\bf Delete this instruction block, but keep the section heading ``NeurIPS Paper Checklist"},
    \item  {\bf Keep the checklist subsection headings, questions/answers and guidelines below.}
    \item {\bf Do not modify the questions and only use the provided macros for your answers}.
\end{itemize}


\begin{enumerate}

\item {\bf Claims}
    \item[] Question: Do the main claims made in the abstract and introduction accurately reflect the paper's contributions and scope?
    \item[] Answer: \answerYes{} 
    \item[] Justification: The abstract has comprehensively reflected the paper's contributions and scope.

\item {\bf Limitations}
    \item[] Question: Does the paper discuss the limitations of the work performed by the authors?
    \item[] Answer: \answerYes{} 
    \item[] Justification: Please refer to the \ref{conclusion} for the limitations of our paper.

\item {\bf Theory assumptions and proofs}
    \item[] Question: For each theoretical result, does the paper provide the full set of assumptions and a complete (and correct) proof?
    \item[] Answer: \answerYes{} 
    \item[] Justification: To prove the frequency-domain structure of the learnable DCT anchors for graph initialization, we provide the theoretical justification in Appendix~\ref{app:dct_proof}.

    \item {\bf Experimental result reproducibility}
    \item[] Question: Does the paper fully disclose all the information needed to reproduce the main experimental results of the paper to the extent that it affects the main claims and/or conclusions of the paper (regardless of whether the code and data are provided or not)?
    \item[] Answer: \answerYes{} 
    \item[] Justification: All the implementation details are provided in Section \ref{app:experimentsettings}.

\item {\bf Open access to data and code}
    \item[] Question: Does the paper provide open access to the data and code, with sufficient instructions to faithfully reproduce the main experimental results, as described in supplemental material?
    \item[] Answer: \answerYes{} 
    \item[] Justification: The source code and dataset are included in the supplementary materials.

\item {\bf Experimental setting/details}
    \item[] Question: Does the paper specify all the training and test details (e.g., data splits, hyperparameters, how they were chosen, type of optimizer, etc.) necessary to understand the results?
    \item[] Answer: \answerYes{} 
    \item[] Justification: All relevant information are provided in Sec~\ref{app:experimentsettings}

\item {\bf Experiment statistical significance}
    \item[] Question: Does the paper report error bars suitably and correctly defined or other appropriate information about the statistical significance of the experiments?
    \item[] Answer: \answerYes{} 
    \item[] Justification: 

\item {\bf Experiments compute resources}
    \item[] Question: For each experiment, does the paper provide sufficient information on the computer resources (type of compute workers, memory, time of execution) needed to reproduce the experiments?
    \item[] Answer: \answerNo{} 
    \item[] Justification: Did not include the computer resources.
    
\item {\bf Code of ethics}
    \item[] Question: Does the research conducted in the paper conform, in every respect, with the NeurIPS Code of Ethics \url{https://neurips.cc/public/EthicsGuidelines}?
    \item[] Answer: \answerYes{} 
    \item[] Justification: NA

\item {\bf Broader impacts}
    \item[] Question: Does the paper discuss both potential positive societal impacts and negative societal impacts of the work performed?
    \item[] Answer: \answerNA{}{} 
    \item[] Justification: Positive impacts include improving DR detection accuracy and supporting clinical workflows.
    
\item {\bf Safeguards}
    \item[] Question: Does the paper describe safeguards that have been put in place for responsible release of data or models that have a high risk for misuse (e.g., pretrained language models, image generators, or scraped datasets)?
    \item[] Answer: \answerNA{} 
    \item[] Justification: Our paper poses no such risks.

\item {\bf Licenses for existing assets}
    \item[] Question: Are the creators or original owners of assets (e.g., code, data, models), used in the paper, properly credited and are the license and terms of use explicitly mentioned and properly respected?
    \item[] Answer: \answerYes{} 
    \item[] Justification: The original paper which produced the MFIDDR dataset is cited in the paper.

\item {\bf New assets}
    \item[] Question: Are new assets introduced in the paper well documented and is the documentation provided alongside the assets?
    \item[] Answer: \answerYes{} 
    \item[] Justification: The code for the experiments are provided in the supplementary materials. The newly generated dataset is provided via an anonymized URL in Appendix~\ref{app:mvgddr}.

\item {\bf Crowdsourcing and research with human subjects}
    \item[] Question: For crowdsourcing experiments and research with human subjects, does the paper include the full text of instructions given to participants and screenshots, if applicable, as well as details about compensation (if any)? 
    \item[] Answer: \answerNA{} 
    \item[] Justification: Our paper does not involve crowdsourcing nor research with human subjects.

\item {\bf Institutional review board (IRB) approvals or equivalent for research with human subjects}
    \item[] Question: Does the paper describe potential risks incurred by study participants, whether such risks were disclosed to the subjects, and whether Institutional Review Board (IRB) approvals (or an equivalent approval/review based on the requirements of your country or institution) were obtained?
    \item[] Answer: \answerNA{} 
    \item[] Justification: Our paper does not involve crowdsourcing nor research with human subjects.

\item {\bf Declaration of LLM usage}
    \item[] Question: Does the paper describe the usage of LLMs if it is an important, original, or non-standard component of the core methods in this research? Note that if the LLM is used only for writing, editing, or formatting purposes and does not impact the core methodology, scientific rigorousness, or originality of the research, declaration is not required.
    \item[] Answer: \answerNA{} 
    \item[] Justification: The core method development in this paper does not involve LLMs as any important, original, or non-standard components.

\end{enumerate}

\end{document}